# Improved Neural Protoform Reconstruction via Reflex Prediction

**Liang Lu, Jingzhi Wang, David R. Mortensen**

Language Technologies Institute, Carnegie Mellon University

{lianglu, jingzhi3, dmortens}@cs.cmu.edu

**Abstract**

Protolanguage reconstruction is central to historical linguistics. The comparative method, one of the most influential theoretical and methodological frameworks in the history of the language sciences, allows linguists to infer protoforms (reconstructed ancestral words) from their reflexes (related modern words) based on the assumption of regular sound change. Not surprisingly, numerous computational linguists have attempted to operationalize comparative reconstruction through various computational models, the most successful of which have been supervised encoder-decoder models, which treat the problem of predicting protoforms given sets of reflexes as a sequence-to-sequence problem. We argue that this framework ignores one of the most important aspects of the comparative method: not only should protoforms be inferable from cognate sets (sets of related reflexes) but the reflexes should also be inferable from the protoforms. Leveraging another line of research—reflex prediction—we propose a system in which candidate protoforms from a reconstruction model are reranked by a reflex prediction model[1]. We show that this more complete implementation of the comparative method allows us to surpass state-of-the-art protoform reconstruction methods on three of four Chinese and Romance datasets.

## 1. Introduction

Historical linguistics provides a window into the human past, the diversification of and interactions between human populations, as well as the mechanisms through which languages change over time. Perhaps the most enduring theoretical and methodological contribution of historical linguistics is the comparative method, by which protolanguages—putative ancestors of families of languages—can be reconstructed (Anttila, 1989; Campbell, 2021). In the comparative method, cognate sets—groups of words believed to have descended from the same ancestral word—are compared in order to infer the corresponding ancestral words (protoforms). These reconstructions are chosen to maximize the regularity of the mapping from reconstructions to reflexes (daughter forms) and minimize the phonetic distance between reconstructions and their reflexes. The assumption is that the historical changes that affect the sounds in words are largely regular such that almost all reflexes in a language can be derived deterministically from the protoforms given a series of sound changes.

This method is difficult to employ in practice, in no small part because datasets can be very large. To deal with the cognitive burden of historical comparison, computational methods have been proposed to assist linguists in this endeavor. However, with a few exceptions (Bouchard-Côté et al., 2013; He et al., 2023; Arora et al., 2023), the most successful comparative reconstruction models have treated this task as a fairly generic sequence-to-sequence transduction task, in essence translating sets of reflexes (cognate sets) into protoforms (Meloni et al., 2021; Chang et al., 2022; Fourrier, 2022; Kim et al., 2023; Cui et al., 2022). This ignores an important aspect of the comparative method in that it does not constrain the protoforms so that they can be deterministically translated back into each of the reflexes.

In this paper, we propose a multi-model reconstruction system that improves its reconstructions via reflex prediction—the task of predicting the reflexes given a protoform. Our system consists of a beam search-enabled sequence-to-sequence reconstruction model and a sequence-to-sequence reflex prediction model that serves as a reranker. The reflex prediction component can often provide valuable information that may not have been captured by the reconstruction model's probability distribution, thereby addressing certain reconstruction errors. We find that our linguistically-motivated method can address some errors made by existing techniques. Figure 1 shows an example where reranking reconstruction candidates according to reflex prediction accuracy compensates for the erroneous probability ranking of the reconstruction model, leading to a correct reconstruction.

As reflex prediction prior art on our datasets of interest is limited, we test various neural reflex prediction models. We then combine pre-trained reconstruction models and reflex prediction models into reconstruction systems. We perform ablation studies and post hoc error analysis to examine the effectiveness of such systems. Our reranked reconstruction system outperforms state-of-the-art neural reconstruction approaches on Meloni et al. (2021)'s Romance datasets and Chang et al. (2022)'s Sinitic dataset WikiHan. Our contributions include:

1. Proposing a multi-model, reranking-driven reconstruction system that achieves state-of-the-art reconstruction results on both Romance and Sinitic datasets
2. Adapting and examining existing architectures, as well as modified variants, for reflex prediction on Romance and Sinitic languages
3. Performing phonologically-informed analysis of the reflex prediction model and its interactions with the reconstruction model in a reranking system
4. Providing a fast implementation of the reconstruction system with vectorized beam search and reranking

---

[1]Our code is available at https://github.com/cmu-llab/reranked-reconstruction.

| Beam Search | | | Reflex Prediction (based on protoform candidates) | | | | | | | | | Reranking Result | | |
|---|---|---|---|---|---|---|---|---|---|---|---|---|---|---|
| rank | $\hat{p}_i^{bs}$ | $m_i$ | | Cantonese | Gan | Hakka | Jin | Mandarin | Hokkien | Wu | Xiang | $r_i$ | rank | $\hat{p}_i^{rk}$ | $s_i$ |
| 0 | pjet入 | -0.1114 | —— | pi:t˧ | pjɛt˥ | pjet˨ | **pjə?˨** | pjɛˇ | pjɛt˧ | **piɪ?˥** | pjɛ˧ | 0.2500 | 0 | **pit入** | 0.5995 |
| 1 | pet入 | -0.2711 | —— | pi:t˧ | pjɛt˥ | pjet˨ | **pjə?˨** | pjɛˇ | pjɛt˧ | **piɪ?˥** | pjɛ˧ | 0.2500 | 1 | pjet入 | 0.2036 |
| 2 | **pit入** | -0.5030 | —— | pet˥ | **pit˥** | **pit˨** | **pjə?˨** | **piˇ** | **pit˧** | **piɪ?˥** | **pi˧** | 0.8750 | 2 | pet入 | 0.0439 |
| 3 | pep入 | -1.5533 | —— | pi:p˧ | pjɛt˥ | pjap˨ | **pjə?˨** | pjɛˇ | pjap˧ | **piɪ?˥** | pjɛ˧ | 0.2500 | 3 | pep入 | -1.2383 |
| 4 | pij去 | -1.6329 | —— | pei˧ | pi˧ | pi˥ | pi˩˥ | piˇ | pi˧ | pi˩˥ | 0.1250 | 4 | pij去 | -1.4754 |
| | | | | **pi:t˥** | **pit˥** | **pit˨** | **pjə?˨** | **piˇ** | **pit˧** | **piɪ?˥** | **pi˧** | - | | | |

Figure 1: A scenario in which using beam search on Meloni et al. (2021)'s sequence-to-sequence GRU reconstruction model incorrectly predicts *pjet*入 as the most likely reconstruction for the 必 *pit*入 'must' cognate set (in the WikiHan test set). The reflex prediction model could only infer 2 (bold) of the 8 reflexes from the incorrect reconstruction *pjet*入, but correctly infers 7 of the 8 reflexes from the third candidate *pit*入. Our reflex prediction-based reranked reconstruction system makes score adjustments that lead to the correct reranked protoform prediction *pit*入. The last row in the reflex prediction table provides reference reflexes. Bold: correct protoform or reflex; *i*: ranking index; $\hat{p}_i^{bs}$: beam search protoform candidate; $m_i$: model score, which is the normalized log probability of the candidate protoform; $r_i$: reranker score; $\hat{p}_i^{rk}$: reranker protoform candidate; $s_i$: adjusted score.

## 2. Related Work

Word form-related tasks in computational historical linguistics include reconstruction, reflex prediction, and cognate prediction[2], as summarized in Figure 2.

### 2.1. Reconstruction

Computational reconstruction of proto-languages was proposed as early as the 1960s (Durham and Rogers, 1969). Bouchard-Côté et al. (2013) used sound change probabilistic models along with a Monte Carlo inference algorithm to automate protoform reconstruction, but their method relies on a phylogenetic tree. List et al. (2022a) proposed sequence comparison and phonetic alignment for reconstruction, but this did not perform well on either the WikiHan or Romance datasets (Cui et al., 2022; Kim et al., 2023). Ciobanu and Dinu (2018) and Ciobanu et al. (2020) used conditional random fields to automate reconstruction by labeling each position in the daughter sequence with a protoform token. Meloni et al. (2021) formulated protoform reconstruction as a sequence-to-sequence task and used an encoder-decoder GRU model to perform Latin reconstruction, setting a baseline for future neural-based reconstruction methods. Fourrier (2022) compared RNNs and Transformers for protoform reconstruction, noting that encoder-decoder architectures can encode phonetic features in the reflexes into an informative latent space from which the decoder can derive the protoform. Extending Meloni et al. (2021)'s work, Kim et al. (2023) proposed using a Transformer-based encoder-decoder architecture with language embedding for protoform reconstruction, achieving state-of-the-art on Meloni et al. (2021)'s Romance dataset and Hóu (2004)'s Sinitic dataset. Very recently, Akavarapu and Bhattacharya (2023) used an MSA Transformer (originally proposed as a protein language model with multiple sequence alignments as inputs (Rao et al., 2021)) pretrained for cognate prediction to perform protoform reconstruction on automatically aligned cognate sets.

---
[2]These terms are sometimes confused. Because we need a distinction here, we categorize them using Arora et al. (2023)'s definitions. When only relatedness but not ancestry is concerned, the protoform is sometimes treated as part of the cognate set in the literature.

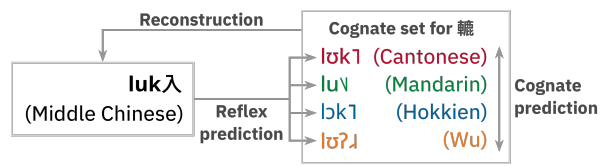

Figure 2: Three word-form-related tasks in historical linguistics, exemplified by the 轆 *luk*入 'wheel' cognate set from WikiHan.

## 2.2. Reflex prediction

Reflex prediction involves modeling the phonological or morphological changes needed to derive reflexes from protoforms. It corresponds to recreating the evolutionary process of languages in historical linguistic studies. Marr and Mortensen (2020, 2023) developed a rule-based Latin-to-French reflex prediction model that predicted reflexes at five different stages in the history of French. Bodt and List (2022) used a semi-automatic method to predict reflexes in Western Kho-Bwa via automatic alignment and identification of sound correspondences on manually annotated cognate sets. Paralleling the use of sequence-to-sequence techniques in reconstruction, Cathcart and Rama (2020) pioneered neural reflex prediction with an LSTM encoder-decoder model that predicts Indo-Aryan languages from Old Indo-Aryan. Recently, Arora et al. (2023) introduced a new South Asian languages dataset and replicated Cathcart and Rama (2020)'s reflex prediction experiments on the dataset with both GRU and Transformer encoder-decoder models. To the best of our knowledge, no work has examined neural reflex prediction with Romance and Sinitic languages.

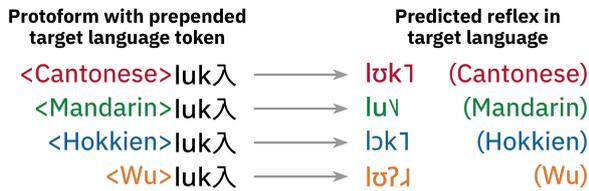

Figure 3: A reflex prediction model aims to derive the correct reflexes based on a protoform sequence tagged by the target daughter languages. A forward pass on the model involves only one daughter language.

## 2.3. Cognate prediction

Nitschke (2021) used neural machine translation techniques to predict missing reflexes in Romance cognate sets. The SIGTYP 2022 shared task on the prediction of conjugate reflexes called for efforts to develop cognate prediction systems and evaluated submissions on numerous language families (List et al., 2022b). Interestingly, a CNN model by Kirov et al. (2022) resembling image-inpainting (Mockingbird-I1) performed the best overall (List et al., 2022b). By treating phonemes as pixels, reflexes as rows of pixels, and cognate sets as stacked rows forming an image, Mockingbird-I1 recovers the missing rows with convolution and deconvolution networks. Cui et al. (2022) found that Mockingbird-I1 can be used to augment a reconstruction dataset, improving the model's stability while training. Although we do not perform cognate prediction in this paper, we test our methods on Cui et al. (2022)'s augmented WikiHan dataset (WikiHan-aug), which will help answer the question of how well reflex prediction, cognate prediction, and reconstruction can combine to form a more effective reconstruction workflow.

## 3. Methods

### 3.1. Datasets

**Romance Datasets:** We use Meloni et al. (2021)'s dataset consisting of both IPA (International Phonetic Alphabet) and orthographic forms. The IPA form (Rom-phon) represents words' pronunciation in phonemes, while the orthographic form (Rom-orth) represents the words as they are spelled out in writing. To compare with the state-of-the-art reconstruction model on the Romance datasets, we match Kim et al. (2023)'s preprocessing and splits.

**Sinitic Datasets:** We use both Hóu (2004)'s dataset compiled by Kim et al. (2023) and Chang et al. (2022)'s WikiHan dataset[3]. Both datasets contain phonetic representations of Middle Chinese and its descendants. We also test an augmented version of WikiHan (WikiHan-aug) created by Cui et al. (2022), which uses the cognate prediction model Mockingbird-I1 to fill in missing daughter entries in the train set. Protoform labels for the Sinitic datasets are based on Baxter (2014)'s reconstructions of Middle Chinese. We match Chang et al. (2022)'s splits for WikiHan and Kim et al. (2023)'s splits for Hóu (2004).

---

[3]Although the full dataset consists of 21,227 cognate sets, cognate sets with less than 3 reflexes are ignored to match previous work.

| Dataset | Cognate sets | # varieties | Ancestor language |
|---|---:|---:|---:|
| WikiHan (Chang et al., 2022) | 5,165 | 8 | Middle Chinese |
| WikiHan-aug (Cui et al., 2022) | 8,780 | 8 | Middle Chinese |
| Hóu (Hóu, 2004) | 804 | 39 | Middle Chinese |
| Rom-phon (Meloni et al., 2021; Ciobanu and Dinu, 2018) | 8,703 | 5 | Latin |
| Rom-orth (Meloni et al., 2021; Ciobanu and Dinu, 2018) | 8,631 | 5 | Latin |

Table 1: Overview of the datasets with their respective number of daughter languages (# varieties) and the number of cognate sets used in the experiments.

### 3.2. Reflex Prediction Models

As there is limited prior work on neural reflex prediction with our datasets of interest, we adapt reflex prediction models previously used for other datasets, along with reconstruction models previously used for the current datasets. Figure 3 shows the reflex prediction task as a sequence-to-sequence transduction task from the protoform—with target language tokens prepended to the beginning—to the reflex in each specified language.

One notable difference between reflex prediction and translation is that, instead of decoding to one target language, the reflex prediction model needs to decode into multiple possible target languages. As a baseline, we attempt various architectural modifications to Meloni et al. (2021)'s unidirectional encoder-decoder GRU model to accommodate multiple target languages—including multi-layer bidirectional encoding, target language embedding during decode similar to Meloni et al. (2021)'s encoder, one-hot vector target language prompting to the decoder's classifier, target-language-specific connections in the decoder's classifier network, and support for VAE-style latent space used by Cui et al. (2022)[4] to decode from the same source to multiple daughters—all of which are tuned as hyperparameters.

Additionally, we adapt Kim et al. (2023)'s Transformer reconstruction model[5] for reflex prediction and test Arora et al. (2023)'s GRU and Transformer reflex prediction models on our datasets of interest[6]. We implement batched training and inference for all the adapted models but retain the original architecture.

---

[4] Cui et al. (2022)'s report proposes a reconstruction model that learns a representation of the cognate set with a Variational Autoencoder (VAE) on the reflexes, reconstructing both the reflexes and the protoforms from the same latent space.

[5] The major difference between Kim et al. (2023)'s model and a standard Transformer encoder-decoder model is the addition of language embeddings for input daughter sequences. In the adapted version for reflex prediction, input language embedding serves little purpose (due to a singular input language) and is thus disabled, making it technically very similar to Arora et al. (2023)'s Transformer model in architecture.

[6] Since Cathcart and Rama (2020)'s model requires additional data such as part of speech for semantic embedding, their reflex prediction model is not fully replicable on our datasets of interest.

### 3.3. Beam Search Reconstruction Model

Since beam search is needed in the reranked reconstruction process and no prior work uses beam search in neural reconstruction, we implement beam search on top of Meloni et al. (2021)'s GRU model (GRU-BS). To isolate the effects of reranking, the architecture of the GRU is kept the same as Meloni et al. (2021), consisting of language and token embeddings, a single-layer unidirectional encoder-decoder GRU model, and a multi-layer perceptron classifier[7]. We tune GRU-BS separately to optimize for performance with beam search.

Before being passed into the reconstruction model, reflexes in a cognate set are concatenated into one long sequence, with separators between the reflexes and language tokens to identify each reflex. An example input is as follows:

`*Cantonese:mei˧*Mandarin:mei˩*Wu:me̞˧*`

### 3.4. Reranked Reconstruction System

In a simple beam search system, the candidate sequences are ranked by their length-normalized log probability, and the candidate with the highest normalized log probability is returned. We propose to enhance this ranking using a reflex prediction model that estimates phonetic naturalness when inferring the reflexes from each candidate protoform, as detailed in Algorithm 1. Given protoform candidates predicted by GRU-BS, we compute the proportion of reflexes correctly derived from each candidate as score adjustment. The candidates are rescored by summing the normalized log probability and the score adjustment, scaled by a score adjustment weight $\lambda$. The candidate with the highest adjusted score is chosen as the final prediction.

### 3.5. Evaluation Criteria

To enable cross-task comparisons, we employ established reconstruction metrics for both reflex prediction and reranked reconstruction experiments, including token edit distance (TED), the number of token insertions, deletions, or substitutions between predictions and targets (Levenshtein et al., 1966); token error rate (TER), a length-normalized edit distance (Cui et al., 2022); accuracy (ACC), the percentage of exactly correct predictions; feature error rate (FER), a measure of phonological edit distance by PanPhon (Mortensen et al., 2016); and B-Cubed F Score (BCFS), a measure of structural similarity between predictions and targets (Amigó et al., 2009; List,

---

[7] We use Chang et al. (2022)'s PyTorch reimplementation obtained from `github.com/cmu-llab/meloni-2021-reimplementation`.

**Algorithm 1** Sequential representation of our reranked reconstruction algorithm
---
**Require:** $d_1, d_2, ..., d_n$ = reflexes in daughter languages $D_1, D_2, ..., D_n$ from a cognate set with $n$ reflexes
**Require:** $f_{\theta_f}$ = a beam search-enabled reconstruction model with pre-trained parameters $\theta_f$
**Require:** $g_{\theta_g}$ = a reflex prediction model with pre-trained parameters $\theta_g$
**Require:** $k$ = beam size for predicting candidate reconstructions on $f_{\theta_f}$
**Require:** $\alpha$ = length normalization constant
**Require:** $\lambda$ = score adjustment weight

$D \leftarrow \text{``*''}D_1\text{``:''}d_1\text{``*''}D_2\text{``:''}d_2\text{``*''}\cdots\text{``*''}D_n\text{``:''}d_n\text{``*''}$ ▷ concatenate reflex sequences into a long sequence, with language labels and separators in between

$C = [(\hat{p}_1, m_1), (\hat{p}_2, m_2), \ldots, (\hat{p}_l, m_l)] \leftarrow f_{\theta_f}(D, k, \alpha)$ ▷ beam search with beam size $k$ to obtain a list of $l \leq k$ candidate protoform predictions $\hat{p}_i$ with their normalized log probabilities $m_i = \frac{\log P(\hat{p}_i|D)}{|\hat{p}_i|^\alpha}$ assigned by $f_{\theta_f}$ for $1 \leq i \leq l$

$C' \leftarrow [\,]$ ▷ initialize reranked candidate list
**for** $(\hat{p}_i, m_i)$ in $C$ **do**
    $a \leftarrow 0$ ▷ counter for the number of correctly derived daughters
    **for** $j \leftarrow 1$ to $n$ **do**
        $\hat{p}'_i \leftarrow D_j\hat{p}_i$ ▷ prepend the $j$-th daughter language token to the candidate protoform
        $\hat{d}_{ij} \leftarrow g_{\theta_g}(\hat{p}'_i)$ ▷ predict the reflex in the $j$-th daughter language based on the $i$-th candidate
        **if** $\hat{d}_{ij} = d_j$ **then**
            $a \leftarrow a + 1$ ▷ increment counter if predicted reflex is correct
    $r_i \leftarrow a/n$ ▷ use the accuracy of reflex predictions as the reranker score $r_i$
    $s_i \leftarrow m_i + \lambda r_i$ ▷ calculate the adjusted score $s_i$ for the $i$-th candidate
    $C' \leftarrow C' \mathbin{+\!\!+} [(\hat{p}_i, s_i)]$ ▷ append entry with adjusted score to reranked candidate list
$C' \leftarrow C'$ sorted by descending $s_i$
**return** $C'[0]$ ▷ return the candidate with the highest adjusted score

---

2019). Tokens are phonemes in all datasets with the exception of the character-level Rom-orth dataset. Consequently, FER cannot be reliably calculated for Rom-orth.

### 3.6. Experiments

**Hyperparameters:** We tune hyperparameters using WandB (Biewald, 2020) except for models already tested by Kim et al. (2023): Meloni et al. (2021)'s GRU reconstruction model and Kim et al. (2023)'s Transformer reconstruction model on Rom-phon, Rom-orth, and Hóu. We use Bayesian search with 100 total runs for the best validation phoneme edit distance, validated every 3 epochs and with early stopping. We keep a constant beam size of 5 when tuning GRU-BS to balance computation cost and effectiveness.

**Reflex Prediction Experiments:** First, we test the reflex prediction capability of the four aforementioned reflex prediction models. For each model on each dataset, we perform 20 runs with random seeds (same hyperparameters). We select the best-performing reflex prediction model from each architecture (GRU or Transformer) as reranker models in reranked reconstruction experiments.

**Baseline Reconstruction Data:** We use both Meloni et al. (2021)'s GRU and Kim et al. (2023) Transformer models as baselines. For datasets present in Kim et al. (2023)'s work, we perform additional runs to obtain 20 runs in total (on top of Kim et al. (2023)'s 10 checkpoints).

**Reranked Reconstruction Experiments:** Each reranking experiment involves a pre-trained GRU-BS reconstruction model and a reflex prediction model acting as a reranker, forming a reranked reconstruction system. We select Arora et al. (2023)'s Transformer and the baseline GRU reflex prediction model as rerankers due to their higher performance among their respective architecture. For each reranked reconstruction system, we tune two additional hyperparameters: the beam size of GRU-BS $k$, and the score adjustment weight $\lambda$. We perform a grid search on the hyperparameters for best validation accuracy[8]. The search results across 20 pairs of pre-trained reconstruction and reflex prediction models are averaged (rounding $k$ to the nearest integer) to obtain the final hyperparameters for evaluations on the test set.

**Statistical Analysis:** Considering a small sample size and unknown distribution, we use Wilcoxon Rank-Sum test (Wilcoxon, 1992) with $\alpha = 0.01$ and Bootstrap test (Efron and Tibshirani, 1994) with 99% confidence interval for the mean difference between models or reconstruction systems.

---
[8]The search range for $k$ and $\lambda$ are [2, 10] with resolution 2 and [0.3, 4.2] with resolution 0.3, respectively.

We consider results to be significant if both tests indicate significance.

**Ablation Studies:** Our reranked reconstruction system extends Meloni et al. (2021)'s model by both beam search and reranking. To isolate the effect of reranking, we remove the reranker to obtain the performance of GRU-BS before reranking (with beam size no larger than when used in reranking) and test for differences in performance between GRU-BS with and without reranking.

**Correlation Experiments:** We select the worst-performing reconstruction model (by accuracy at $k = 5$) and rerank it with all the pre-trained reflex prediction models (20 GRU and 20 Transformer) using the same reranking hyperparameters obtained from grid search, effectively varying the reranker with controlled reconstruction and reranking hyperparameters. We then examine the correlation between reflex prediction performance and reranked reconstruction performance.

## 4. Results and Discussion

### 4.1. Reflex prediction Results

Table 2 shows the average performance of the four reflex prediction models. We found statistically significant evidence that Arora et al. (2023)'s Transformer performs the best on all metrics for WikiHan and WikiHan-aug, Arora et al. (2023)'s Transformer performs the best only on ACC for Rom-orth, and Kim et al. (2023)'s Transformer performs the best on TER, TED, and BCFS for Rom-phon. We find no evidence that the top-2 performing models are statistically different for the remaining metrics or datasets. Among GRU models, the baseline performs better than Arora et al. (2023)'s GRU model across all datasets.

On all datasets except Rom-phon, we obtain overall better performance at reflex prediction than reconstruction—consistent with the hypothesis that learning regular sound changes is easier in the forward direction. It is possible that reflex prediction performs worse than reconstruction on Rom-orth due to its non-phonetic nature, obscuring the environments for sound changes.

In reranking experiments, we select the best-performing model for each architecture: Arora et al. (2023)'s Transformer and the baseline GRU[9].

### 4.2. Reranked Reconstruction Results

As shown in Table 3, our reranking system performs significantly better than both Meloni et al. (2021) and Kim et al. (2023) on all datasets except Hóu, for which it performs better on some metrics. We notice a high variance in both reflex prediction and reconstruction results for Hóu, possibly due to its small test set. Finally, we find no statistical difference between using a GRU or a Transformer as a reranker, despite evidence that Transformers outperform GRUs on reflex prediction.

Cui et al. (2022) previously found no evidence that data augmentation helps improve reconstruction. However, our result on WikiHan-aug suggests that cognate set augmentation contributes to both reflex prediction and reranked reconstruc-tion performance, bringing WikiHan reconstruction accuracy to 3% above previous work.

---
[9] We also tested reranking using Kim et al. (2023)'s Transformer model, but found no statistical difference in performance.

| Dataset | Model | ACC%↑ | TED↓ | TER↓ | FER↓ | BCFS↑ |
|---|---|---|---|---|---|---|
| WikiHan | GRU (baseline) | 66.43% | 0.5244 | 0.1547 | 0.0400 | 0.7394 |
| | GRU (Arora et al., 2023) | 64.45% | 0.5558 | 0.1640 | 0.0428 | 0.7260 |
| | Transformer (Kim et al., 2023) | 66.39% | 0.5302 | 0.1564 | 0.0406 | 0.7370 |
| | Transformer (Arora et al., 2023) | **67.64%** | **0.5128** | **0.1513** | **0.0390** | **0.7445** |
| WikiHan-aug | GRU (baseline) | 68.11% | 0.5007 | 0.1477 | 0.0380 | 0.7495 |
| | GRU (Arora et al., 2023) | 66.94% | 0.5159 | 0.1522 | 0.0391 | 0.7430 |
| | Transformer (Kim et al., 2023) | 68.96% | 0.4889 | 0.1442 | 0.0371 | 0.7551 |
| | Transformer (Arora et al., 2023) | **69.37%** | **0.4826** | **0.1424** | **0.0363** | **0.7572** |
| Hóu | GRU (baseline) | 51.72% | 0.7777 | 0.2037 | 0.0488 | 0.6783 |
| | GRU (Arora et al., 2023) | 49.26% | 0.8266 | 0.2166 | 0.0528 | 0.6622 |
| | Transformer (Kim et al., 2023) | 55.46% | 0.7576 | 0.1985 | 0.0494 | 0.6882 |
| | Transformer (Arora et al., 2023) | **55.60%** | **0.7520** | **0.1970** | **0.0485** | **0.6892** |
| Rom-phon | GRU (baseline) | 63.85% | 0.7439 | 0.1014 | **0.0426** | 0.8361 |
| | GRU (Arora et al., 2023) | 48.28% | 1.3257 | 0.1808 | 0.0930 | 0.7567 |
| | Transformer (Kim et al., 2023) | **64.19%** | **0.7349** | **0.1002** | 0.0427 | **0.8380** |
| | Transformer (Arora et al., 2023) | 63.96% | 0.7442 | 0.1015 | 0.0428 | 0.8361 |
| Rom-orth | GRU (baseline) | 64.58% | 0.7301 | 0.0967 | - | 0.8465 |
| | GRU (Arora et al., 2023) | 57.92% | 0.8741 | 0.1158 | - | 0.8218 |
| | Transformer (Kim et al., 2023) | 64.80% | 0.7258 | 0.0961 | - | **0.8478** |
| | Transformer (Arora et al., 2023) | **65.20%** | **0.7247** | **0.0960** | - | 0.8476 |

Table 2: Average performance of the reflex prediction models across 20 runs, with bold indicating the best-performing model for each metric.

| Dataset | Reconstruction System | ACC%↑ | TED↓ | TER↓ | FER↓ | BCFS↑ |
|---|---|---|---|---|---|---|
| WikiHan | GRU (Meloni et al., 2021) | 55.58% | 0.7360 | 0.1724 | 0.0686 | 0.7426 |
| | Trans (Kim et al., 2023) | 54.62% | 0.7453 | 0.1746 | 0.0696 | 0.7393 |
| | GRU-BS ($k = 10$) | 54.88% | 0.7507 | 0.1758 | 0.0701 | 0.7364 |
| | GRU-BS ($k \leq 10$) + GRU Reranker | 57.14%*† | 0.7045*† | 0.1650*† | 0.0661*† | 0.7515*† |
| | GRU-BS ($k \leq 10$) + Trans. Reranker | **57.26%*†** | **0.7029*†** | **0.1646*†** | **0.0658*†** | **0.7520*†** |
| WikiHan-aug | GRU (Meloni et al., 2021) | 54.73% | 0.7574 | 0.1774 | 0.0689 | 0.7346 |
| | Trans (Kim et al., 2023) | 55.82% | 0.7317 | 0.1714 | 0.0661 | 0.7416 |
| | GRU-BS ($k = 10$) | 56.64%* | 0.7214 | 0.1690 | 0.0658 | 0.7454 |
| | GRU-BS ($k \leq 10$) + GRU Reranker | **58.58%*†** | **0.6822*†** | **0.1598*†** | 0.0628*† | **0.7579*†** |
| | GRU-BS ($k \leq 10$) + Trans. Reranker | 58.58%*† | 0.6840*† | 0.1602*† | **0.0626*†** | 0.7575*† |
| Hóu | GRU (Meloni et al., 2021) | 34.63% | 1.0916 | 0.2479 | 0.0914 | 0.6697 |
| | Trans (Kim et al., 2023) | 39.01% | 0.9904 | 0.2233 | 0.0875 | 0.6955 |
| | GRU-BS ($k = 10$) | 37.36% | 1.0382 | 0.2328 | 0.0917 | 0.6974 |
| | GRU-BS ($k \leq 10$) + GRU Reranker | 40.50%† | 0.9727† | 0.2181† | 0.0867† | 0.7130*† |
| | GRU-BS ($k \leq 10$) + Trans. Reranker | **42.08%*†** | **0.9503*†** | **0.2131*†** | **0.0850†** | **0.7170*†** |
| Rom-phon | GRU (Meloni et al., 2021) | 51.92% | 0.9775 | 0.1244 | 0.0390 | 0.8275 |
| | Trans (Kim et al., 2023) | 53.04% | 0.9050 | 0.1148 | 0.0377 | 0.8417 |
| | GRU-BS ($k = 10$) | 52.63% | 0.9125 | 0.1018* | 0.0353* | 0.8402 |
| | GRU-BS ($k \leq 10$) + GRU Reranker | **53.95%*†** | 0.8775*† | 0.0979*† | 0.0336*† | 0.8460*† |
| | GRU-BS ($k \leq 10$) + Trans. Reranker | 53.85%*† | **0.8765*†** | **0.0978*†** | **0.0333*†** | **0.8461*†** |
| Rom-orth | GRU (Meloni et al., 2021) | 69.41% | 0.6004 | 0.0781 | - | 0.8906 |
| | Trans (Kim et al., 2023) | 71.05% | 0.5636 | 0.0734 | - | 0.8981 |
| | GRU-BS ($k = 10$) | 71.09% | 0.5531 | 0.0617* | - | 0.8990 |
| | GRU-BS ($k \leq 10$) + GRU Reranker | **72.60%*†** | **0.5237*†** | **0.0584*†** | - | **0.9045*†** |
| | GRU-BS ($k \leq 10$) + Trans. Reranker | 72.50%*† | 0.5246*† | 0.0585*† | - | 0.9044*† |

Table 3: Evaluation of reconstruction systems, including baselines, GRU with beam search (GRU-BS), and GRU-BS with reranking, averaged across 20 runs. Bold indicates the best-performing system for each metric, asterisks indicate statistically better performance than both baseline models (Meloni et al. (2021)'s GRU and Kim et al. (2023)'s Transformer), and daggers indicate that a reranking system performs statistically better than its beam search counterpart.

| Dataset | Reranker | ACC | TED | TER | FER | BCFS |
|---|---|---|---|---|---|---|
| WikiHan | GRU Reranker | 0.2771 | 0.3647 | 0.3647 | 0.1845 | 0.3639 |
| | Trans. Reranker | 0.3860 | 0.0987 | 0.0987 | 0.3132 | 0.0276 |
| WikiHan-aug | GRU Reranker | 0.3830 | 0.4179 | 0.4179 | 0.4829 | 0.3573 |
| | Trans. Reranker | 0.1849 | -0.0435 | -0.0435 | -0.0655 | -0.0310 |
| Hóu | GRU Reranker | 0.3735 | 0.1236 | 0.1236 | 0.4278 | 0.0173 |
| | Trans. Reranker | 0.5432 | 0.2742 | 0.2742 | 0.2782 | 0.3184 |
| Rom-phon | GRU Reranker | -0.2207 | -0.0115 | -0.0115 | -0.0373 | 0.0336 |
| | Trans. Reranker | -0.0706 | -0.0181 | -0.0181 | -0.0866 | 0.0421 |
| Rom-orth | GRU Reranker | 0.2531 | 0.4044 | 0.4044 | - | 0.4459 |
| | Trans. Reranker | 0.1639 | 0.1123 | 0.1123 | - | 0.1035 |

Table 4: Correlation coefficients between rerankers' reflex prediction performance and reranked reconstruction performance. The cells are color-coded by sign and strength, with red for positive correlation coefficients and blue for negative correlation coefficients.

### 4.3. Correlation Test Results

Correlation analysis reveals a mostly positive correlation between rerankers' reflex prediction performance and the corresponding reranking system's reconstruction performance, except on the Rom-phon dataset (see Table 4). Although statistical significance is unclear due to small sample sizes, evidence suggests that the performance of the reranker could play an important role in a reranked reconstruction system.

### 4.4. Ablation Studies

While GRU-BS alone (without a reranker) outperforms baseline models on some occasions, GRU-BS with a reranker performs statistically better than GRU-BS alone for all datasets and metrics, as indicated in Table 3. Even though beam search is commonly regarded as a powerful method in sequence-to-sequence tasks, its ability in a pro-tolanguage reconstruction setting is still limited compared to reranking, where modeling reflex prediction in addition to reconstruction proves more informative.

### 4.5. Reranking Error Analysis

To gain insights into the reranker's behavior, we conduct error analyses on the top-performing reranking system (GRU-BS + Transformer Reranker) by randomly selecting one of the 20 runs. We denote the ranks from beam search and reranking by $r_{bs}$ and $r_{rk}$ respectively (better score has lower rank), and categorized the reranker's behavior into four distinct categories:

- **Improved** ($r_{rk} < r_{bs}$): reranker assigns a more favorable rank to the target protoform.
- **Unchanged** ($r_{rk} = r_{bs}$): reranker does not alter the rank of the target protoform.
- **Worsened** ($r_{rk} > r_{bs}$): reranker assigns a less desirable rank to the target protoform.
- **Not-in**: the target protoform is not predicted as a candidate by beam search and is thus not seen by the reranker. This category is not included in analyses that require the target protoform to be processed by the reranker.

Table 5 shows the distribution of the reranker's behavior among the four categories. On every dataset, the reranker improves the ranking of the target protoform more often than worsens it.

We observe that, compared to the target protoform, the incorrectly predicted protoform often exhibits greater phonetic similarity, measured by both token edit distance and feature edit distance, to the reflexes[10] (see Table 6). It is likely that the reflex prediction models find it easier to derive correct reflexes from predicted protoforms that are more similar to the reflexes, potentially making it challenging for the reranker to improve the ranking of the target protoforms less similar to the reflexes.

Furthermore, certain sound combinations in WikiHan's Middle Chinese forms, such as *ju*, *je*, and *xwo*, are absent in the daughter languages included in the dataset (see Table 7 for some examples). This highlights a notable challenge of computational reconstruction—recovering phonemes lost during language evolution—which likely require solutions other than reranking.

Finally, we observe that the reranker model has the highest overall error rate when predicting Hokkien reflexes compared to other daughter languages on WikiHan (see Table 8), despite Hokkien having the third most training examples. A possible explanation is Karlgren (1974)'s hypothetical subgrouping of Sinitic in which Hokkien is a descendent of a sister of Middle Chinese rather than Middle Chinese itself.

---

[10]Because the phonetic values of Middle Chinese tones are unknown (WikiHan represents them tones with the four abstract tone characters from Tang Dynasty rhyme books), we exclude tones when calculating $D_T$ and $D_F$ in Table 6 and in the case studies in Table 7.

| Dataset | Improved | Worsened | Unchanged | Not-in | Total | Improved/Changed (%) |
|---|---|---|---|---|---|---|
| WikiHan | 84 (8.13%) | 32 (3.10%) | 755 (73.09%) | 162 (15.68%) | 1033 | 72.41% |
| WikiHan-aug | 88 (8.52%) | 23 (2.23%) | 791 (76.57%) | 131 (12.68%) | 1033 | 79.28% |
| Hóu | 26 (16.15%) | 15 (9.32%) | 88 (54.66%) | 32 (19.88%) | 161 | 63.41% |
| Rom-phon | 109 (6.21%) | 61 (3.48%) | 1198 (68.30%) | 386 (22.01%) | 1754 | 64.12% |
| Rom-orth | 75 (4.29%) | 23 (1.32%) | 1367 (78.16%) | 284 (16.24%) | 1749 | 76.53% |

Table 5: The distribution of reranker behavior categorization on the test set (left), based on a randomly sampled run for each dataset, as well as the corresponding rate of ranking improvement among instances with changed (i.e. improved or worsened) ranking (right).

| Dataset | Category | $D_T(\hat{p}, R) < D_T(p, R)$ (R more similar to $\hat{p}$ than $p$ by $D_T$) | $D_F(\hat{p}, R) < D_F(p, R)$ (R more similar to $\hat{p}$ than $p$ by $D_F$) |
|---|---|---|---|
| WikiHan | Worsened | **37.50%** | **53.12%** |
| | Unchanged | 35.56% | 43.33% |
| | Improved | 28.30% | 32.08% |
| Rom-phon | Worsened | **60.66%** | **62.30%** |
| | Unchanged | 47.21% | 51.48% |
| | Improved | 47.17% | 49.06% |

Table 6: Comparison between the phonetic similarity between the reflexes $R$ and the predicted protoform $\hat{p}$ versus the target protoform $p$ for each category of the reranker's behavior among reconstruction errors. Similarity are measured by normalized token edit distance ($D_T$) and normalized feature edit distance ($D_F$). The table presents percentages of entries in each category where the predicted protoforms exhibit greater phonetic similarity to their modern reflexes than the target protoforms according to each similarity metric, with bold indicating the highest percentage.

| Dataset | Category | Worsened | | Unchanged | | Improved | |
|---|---|---|---|---|---|---|---|
| | | Proto | Prôto | Proto | Prôto | Proto | Prôto |
| WikiHan | Middle Chinese | **mjukʷ** | **mukʷ** | **tʰ ja ŋ** | **tɕʰa ŋ** | **tʃʰje k** | **tʃ e k** |
| | Cantonese | m⌴ʊk | mʊk | tʃʰ⌴ɔːŋ | tʃʰ⌴ɔːŋ | tʃ⌴ɪ k | tʃ ɪ k |
| | Hakka | m⌴uk | muk | | | tʃ⌴a k | tʃ a k |
| | Mandarin | m⌴u⌴ | mu⌴ | tʃʰ⌴a ŋ | tʃʰa ŋ | tɕʰ⌴i ⌴ | tɕʰi ⌴ |
| | Hokkien | b⌴ɔk | bɔk | tɕʰ⌴iɔŋ | tɕʰiɔŋ | tɕʰ⌴iək | tɕʰiək |
| Rom-phon | Latin | **astʰma** | **as⌴ma** | **f⌴ɛrɪtatɛm** | **f⌴ɛritam** | **tɛksɛrɛ** | **tɪssɛrɛ** |
| | Romanian | ast mə | as<u>t</u>mə | | | t⌴⌴sese | t⌴⌴sese |
| | French | as⌴ m⌴ | as⌴m⌴ | fjɛʁ⌴te⌴⌴⌴ | fjɛʁ⌴te⌴ | ti⌴se⌴⌴ | ti⌴se⌴⌴ |
| | Italian | az⌴ ma | az⌴ma | f⌴erita⌴⌴⌴ | f⌴erita⌴ | tɛssere | tɛssere |
| | Spanish | as⌴ ma | as⌴ma | | | tex⌴er⌴ | tex⌴er⌴ |
| | Portuguese | aʒ⌴ me | aʒ⌴me | | | ti⌴seɪ⌴ | ti⌴seɪ⌴ |

Color key: ■ substitution  ■ insertion  ■ (⌴) deletion

Table 7: Instances in each category where the predicted protoform is phonetically closer to its reflexes than the target protoform by both $D_T$ and $D_F$, selected from the WikiHan (top) and Rom-phon (bottom) test sets. The **Proto** and **Prôto** columns show edits from the target protoform and the predicted protoform to the reflexes, respectively. Words in each column are manually aligned to reflect edits, with '⌴' indicating an empty position in the multi-sequence alignment in the case of deletion or insertion. Unavailable reflexes are not shown on the table, and languages without available reflexes (Gan, Jin, Wu, and Xiang) are omitted. Differences in edits between the predicted and target protoforms are shaded. The selected entries are 睦 *mjukʷ* 'friendly', 昶 *tʰjaŋ* 'long daytime', 磧 *tʃʰjek* 'gravel' (WikiHan), *asthma* 'asthma', *feritatem* 'ferocity', and *texere* 'to weave' (Rom-phon).

| Category | Cantonese | Gan | Hakka | Jin | Mandarin | Hokkien | Wu | Xiang |
|---|---|---|---|---|---|---|---|---|
| Not-in | 75.31% | 72.73% | 70.83% | 83.33% | 72.22% | 82.05% | 57.69% | 82.14% |
| Worsened | 59.38% | 75.00% | 40.00% | 71.43% | 68.75% | 68.75% | 33.33% | 60.00% |
| Unchanged | 24.44% | 21.21% | 42.16% | 15.79% | 27.22% | 46.86% | 15.66% | 41.67% |
| Improved | 38.46% | 25.00% | 54.05% | 13.33% | 18.87% | 50.00% | 37.93% | 41.18% |
| Overall | 48.12% | 40.00% | 53.10% | 42.22% | 46.37% | 62.29% | 32.95% | 55.81% |

Table 8: Reranker's reflex prediction error rates among reranked reconstruction error entries (when predicting reflexes from the target protoform) for each daughter language in the WikiHan dataset given each reranker behavior category, obtained from a randomly selected run.

## 5. Conclusion

Ironically, many efforts to automate protolanguage reconstruction with neural models have thus far treated reconstruction as a sequence-to-sequence task, disregarding the comparative method's constraint that reflexes should be inferable from the reconstructions. Our reranked reconstruction system provides an elegant way to replicate the synergy between reconstruction and reflex prediction in the comparative method, yielding results that surpass existing methods—a vindication of the idea that designing reconstruction systems with the comparative method in mind can be more powerful than relying solely on sequence-to-sequence techniques.

Though our approach yields better reconstruction performance, it is left to future work to address some of the challenges identified in the present work, such as a reconstruction system's tendency to produce reconstructions relatively similar to the reflexes. In the bigger picture, reranking is but one way to bring together multiple tasks in historical linguistics, and arguably a complicated one due to its multi-step training and tuning process. Future research, therefore, could also explore approaches to integrate reconstruction and reflex prediction into one seamless model.

## 6. Acknowledgements

This work is supported by Carnegie Mellon University's SURF grant. We thank Kalvin Chang and Chenxuan Cui for ideas and discussions that informed this work, Kalvin Chang for guidance on using their preprocessed reconstruction datasets and their PyTorch reimplementation of Meloni et al. (2021), Aryaman Arora for guidance regarding their reflex prediction models, and Kalvin Chang, Anna Cai, and Ting Chen for help with revision and proofreading.

## 7. Bibliographical References

## 8. Language Resource References

## A. Hyperparameters

We tune hyperparameters for all models on each dataset using WandB (Biewald, 2020) except for those tested by Kim et al. (2023) (which includes Meloni et al. (2021)'s GRU reconstruction model and Transformer reconstruction model on Rom-phon, Rom-orth, and Hóu). We use Bayesian search with 100 total runs (with early stopping) for the best validation phoneme edit distance (validated every 3 epochs). We keep a constant beam size of 5 for GRU-BS reconstruction models during tuning. Tables 10, 11, 14, 13, 15, and 16 report our hyperparameter search results[11].

Adam optimizer's $\beta_1 = 0.9$, $\beta_2 = 0.999$, and $\varepsilon = 1e{-}8$ are obtained from Chang et al. (2022)'s experiments, while $\beta_1 = 0.9$, $\beta_2 = 0.98$, and $\varepsilon = 1e{-}9$ are used to consistently replicate Arora et al. (2023)'s experiments. We do not observe a noticeable effect $\beta_2$ and $\varepsilon$ have on the models' performance.

## B. Dataset Source and Splits

The WikiHan dataset can be obtained from Chang et al. (2022), and Hóu (2004)'s dataset can be obtained through Kim et al. (2023). WikiHan-aug is obtained from Cui et al. (2022). The Romance datasets by Meloni et al. (2021) is not licensed for redistribution and thus not included in our repository.

All the datasets are split by 70%, 10%, and 20% into train, validation, and test sets. The splits for WikiHan Chang et al. (2022) match the original work, and the splits for Meloni et al. (2021) and Hóu (2004) match Kim et al. (2023). WikiHan-aug Cui et al. (2022) has the same validation and test sets as Chang et al. (2022) but with augmented reflexes in the train set. Because Chang et al. (2022) only included cognate sets with at least 3 daughters in the train set, the train set in WikiHan-aug includes additional cognate sets that fulfill the 3-daughter requirement after augmentation.

Daughter languages included in WikiHan are Cantonese, Gan, Hakka, Jin, Mandarin, Hokkien, Wu, and Xiang. Daughter languages included in Rom-phon and Rom-orth are French, Italian, Spanish, Romanian, and Portuguese. Daughter languages included in Hóu are Beijing, Harbin, Tianjin, Jinan, Qingdao, Zhengzhou, Xian, Xining, Yinchuan, Lanzhou, Urumqi, Wuhan, Chengdu, Guiyang, Kunming, Nanjing, Hefei, Taiyuan, Pingyao, Hohhot, Shanghai, Suzhou, Hangzhou, Wenzhou, Shexian, Tunxi, Changsha, Xiangtan, Nanchang, Meixian, Taoyuan, Guangzhou, Nanning, Hong Kong, Xiamen, Fuzhou, Jianou, Shantou, and Haikou.

## C. Training

All models are trained on NVIDIA GeForce RTX 2080 Ti or RTX A6000 GPUs. Each run takes about 1–3 hours of compute time. Our total GPU compute time is 237 days.

---

[11]We use batch size to refer to the number of cognate sets in a batch, meaning that the number of reflex prediction training examples in each batch may vary if cognate sets have missing daughters.

## D. Reranking Hyperparameters

The optimal beam size $k$ and score adjustment weight $\lambda$ can be dataset-dependent. We use grid search with ranges and resolutions detailed in Table 9 to optimize $k$ and $\lambda$ on the validation set. The search results are shown in Table 17. We observe a preference for higher $\lambda$ on Sinitic datasets.

| Hyperparameter | Search range (inclusive) | Resolution |
|---|---|---|
| Beam size $k$ | [2, 10] | 2 |
| Score adjustment weight $\lambda$ | [0.3, 4.2] | 0.3 |

Table 9: Grid search range and resolution for reranking hyperparameters.

## E. Results with standard deviation

Table 18 shows reflex prediction performance with standard deviations, and Table 19 shows reconstruction performance with standard deviations.

## F. Additional Reflex Error Analysis

Table 20 shows the reflex prediction error rate for each daughter among all test entries. Similar to Table 8, we observe an overall highest error rate on Hokkien.

## G. Statistical Tests Results

We obtain *p*-values from the Wilcoxon Rank-Sum test and confidence intervals (CI) from the Bootstrap test. Tables 21, 22, 23, 24, and 25 show *p*-values and 99% confidence intervals for reflex prediction performance. Tables 26, 27, 28, 29, and 30 show *p*-values and 99% confidence intervals for reconstruction performance.

## H. Additional Reranking Case Studies

We provide additional reranking examples similar to Figure 1. Figures 4 and 5 show two additional reranking successes on WikiHan, Figures 6 and 7 show two failures on WikiHan, Figures 8 and 9 show two successes on Rom-phon, and Figures 10 and 11 show two failures on Rom-phon.

|                | WikiHan     | WikiHan-aug | Hóu        | Rom-phon   | Rom-orth   |
|----------------|-------------|-------------|------------|------------|------------|
| batch size $k$ | 128         | 256         | 32         | 256        | 256        |
| beam search $a$| 0.912598    | 0.600524    | 0.638660   | 0.825868   | 0.707860   |
| dropout        | 0.405044    | 0.496428    | 0.497715   | 0.430556   | 0.489005   |
| embedding size | 509         | 148         | 265        | 154        | 283        |
| feedforward size | 218       | 471         | 232        | 310        | 311        |
| hidden size    | 81          | 216         | 36         | 115        | 255        |
| number of layers | 1         | 1           | 1          | 1          | 1          |
| learning rate  | 0.000629980 | 0.000550343 | 0.000691970| 0.000762067| 0.000568855|
| max epochs     | 576         | 204         | 194        | 285        | 304        |
| warmup epochs  | 19          | 3           | 24         | 50         | 50         |
| $\beta_1$      | 0.9         | 0.9         | 0.9        | 0.9        | 0.9        |
| $\beta_2$      | 0.999       | 0.999       | 0.999      | 0.999      | 0.999      |
| $\varepsilon$  | 1e–8        | 1e–8        | 1e–8       | 1e–8       | 1e–8       |

Table 10: Hyperparameters for GRU reconstruction model with beam search (GRU-BS), tuned with fixed beam size $k = 5$. Beam search $a$ is the length normalization constant. The number of GRU layers is set to 1 to match Meloni et al. (2021).

|                                 | WikiHan     | WikiHan-aug | Hóu        | Rom-phon   | Rom-orth   |
|---------------------------------|-------------|-------------|------------|------------|------------|
| batch size                      | 128         | 256         | 128        | 64         | 256        |
| learning rate                   | 0.000610810 | 0.00128592  | 0.00208360 | 0.000153890| 0.000931776|
| max epochs                      | 280         | 202         | 485        | 487        | 371        |
| dropout                         | 0.422406    | 0.411611    | 0.402412   | 0.467993   | 0.481404   |
| embedding size                  | 328         | 286         | 46         | 324        | 41         |
| feedforward size                | 421         | 183         | 500        | 275        | 96         |
| target-gated classifier         | True        | False       | False      | False      | True       |
| decode with language embedding  | False       | False       | True       | False      | False      |
| hidden size                     | 46          | 33          | 110        | 177        | 194        |
| number of encoder layers        | 2           | 4           | 1          | 2          | 1          |
| one-hot target encoding         | True        | True        | True       | True       | False      |
| bidirectional encoder           | True        | True        | True       | True       | True       |
| use VAE latent                  | False       | True        | False      | False      | False      |
| warmup epochs                   | 0           | 42          | 28         | 41         | 6          |
| $\beta_1$                       | 0.9         | 0.9         | 0.9        | 0.9        | 0.9        |
| $\beta_2$                       | 0.999       | 0.999       | 0.999      | 0.999      | 0.999      |
| $\varepsilon$                   | 1e–8        | 1e–8        | 1e–8       | 1e–8       | 1e–8       |

Table 11: Hyperparameters for our GRU reflex prediction model. Target-gated classifier refers to whether each specific target language enables a specific subset of the token classifier, decode with language embedding refers to whether to embed the target sequences with language embedding similar to Meloni et al. (2021), one-hot target encoding refers to whether the classifier is prompted with an additional one-hot vector concatenated to its input to indicate the target daughter language, and use VAE latent refers to whether the decoder takes a sampled and reparametrized latent similar to Cui et al. (2022)'s VAE reconstruction model.

|                          | WikiHan     | WikiHan-aug | Hóu        | Rom-phon   | Rom-orth   |
|--------------------------|-------------|-------------|------------|------------|------------|
| batch size               | 512         | 32          | 32         | 128        | 64         |
| learning rate            | 0.00133100  | 0.000275041 | 0.00223748 | 0.00103299 | 0.00117782 |
| max epochs               | 413         | 186         | 177        | 514        | 383        |
| dropout                  | 0.109371    | 0.352477    | 0.239702   | 0.159863   | 0.250876   |
| embedding size           | 128         | 128         | 64         | 64         | 64         |
| feedforward size         | 429         | 1002        | 962        | 275        | 677        |
| nhead                    | 1           | 16          | 16         | 2          | 8          |
| number of decoder layers | 2           | 8           | 7          | 3          | 2          |
| number of encoder layers | 5           | 2           | 4          | 6          | 3          |
| warmup epochs            | 20          | 17          | 40         | 5          | 37         |
| weight_decay             | 6.29736e-07 | 5.34183e-07 | 9.50859e-07| 4.76354e-07| 9.89944e-07|
| $\beta_1$                | 0.9         | 0.9         | 0.9        | 0.9        | 0.9        |
| $\beta_2$                | 0.999       | 0.999       | 0.999      | 0.999      | 0.999      |
| $\varepsilon$            | 1e–8        | 1e–8        | 1e–8       | 1e–8       | 1e–8       |

Table 12: Hyperparameter for Kim et al. (2023)'s Transformer model adapted for reflex prediction.

|  | **WikiHan** | **WikiHan-aug** | **Hóu** | **Rom-phon** | **Rom-orth** |
|---|---|---|---|---|---|
| batch size | 256 | 512 | 32 | 256 | 32 |
| bidirectional encoder | True | True | True | True | True |
| dropout | 0.380055 | 0.434051 | 0.170343 | 0.278587 | 0.337712 |
| embedding size | 319 | 284 | 169 | 508 | 473 |
| hidden size | 353 | 397 | 367 | 448 | 422 |
| learning rate | 0.000286969 | 0.000321132 | 0.00143399 | 0.000153972 | 0.00146615 |
| max epochs | 506 | 434 | 542 | 436 | 179 |
| number of layers | 2 | 2 | 4 | 4 | 2 |
| warmup epochs | 37 | 43 | 14 | 16 | 10 |
| $\beta_1$ | 0.9 | 0.9 | 0.9 | 0.9 | 0.9 |
| $\beta_2$ | 0.98 | 0.98 | 0.98 | 0.98 | 0.98 |
| $\varepsilon$ | 1e–9 | 1e–9 | 1e–9 | 1e–9 | 1e–9 |

Table 13: Hyperparameters for Arora et al. (2023)'s GRU reflex prediction model tuned on our datasets of interest.

|  | **WikiHan** | **WikiHan-aug** | **Hóu** | **Rom-phon** | **Rom-orth** |
|---|---|---|---|---|---|
| batch size | 32 | 256 | 32 | 64 | 32 |
| feedforward size | 295 | 832 | 535 | 786 | 110 |
| model size | 64 | 64 | 256 | 128 | 64 |
| dropout | 0.199505 | 0.447951 | 0.443183 | 0.263578 | 0.251537 |
| learning rate | 0.000561076 | 0.00234191 | 0.00260176 | 0.00161177 | 0.00164335 |
| max epochs | 361 | 437 | 536 | 207 | 264 |
| nhead | 4 | 2 | 4 | 1 | 4 |
| number of layers | 4 | 5 | 3 | 5 | 7 |
| warmup epochs | 9 | 49 | 11 | 2 | 9 |
| $\beta_1$ | 0.9 | 0.9 | 0.9 | 0.9 | 0.9 |
| $\beta_2$ | 0.98 | 0.98 | 0.98 | 0.98 | 0.98 |
| $\varepsilon$ | 1e–9 | 1e–9 | 1e–9 | 1e–9 | 1e–9 |

Table 14: Hyperparameters for Arora et al. (2023)'s Transformer reflex prediction model tuned on our datasets of interest.

|  | **WikiHan** | **WikiHan-aug** |
|---|---|---|
| batch size | 512 | 256 |
| dropout | 0.431211 | 0.409475 |
| embedding size | 248 | 196 |
| feedforward size | 375 | 421 |
| decode with language embedding | True | True |
| hidden size | 78 | 278 |
| number of layers | 1 | 1 |
| bidirectional encoder | False | False |
| use VAE latent | False | False |
| learning rate | 0.000935879 | 0.000964557 |
| max epochs | 472 | 298 |
| warmup epochs | 18 | 26 |
| $\beta_1$ | 0.9 | 0.9 |
| $\beta_2$ | 0.999 | 0.999 |
| $\varepsilon$ | 1e–8 | 1e–8 |

Table 15: Hyperparameters for Meloni et al. (2021)'s GRU reconstruction model on WikiHan and WikiHan-aug. For the hyperparameters used to train the same model on Hóu, Rom-phon, and Rom-orth, refer to Kim et al. (2023).

|  | **WikiHan** | **WikiHan-aug** |
|---|---|---|
| batch size | 512 | 64 |
| dropout | 0.170582 | 0.293413 |
| embedding size | 256 | 64 |
| feedforward size | 133 | 857 |
| nhead | 1 | 2 |
| number of decoder layers | 5 | 3 |
| number of encoder layers | 4 | 6 |
| learning rate | 0.000556150 | 0.000595262 |
| max epochs | 194 | 209 |
| warmup epochs | 1 | 4 |
| weight decay | 8.48140e-07 | 8.26112e-07 |
| $\beta_1$ | 0.9 | 0.9 |
| $\beta_2$ | 0.999 | 0.999 |
| $\varepsilon$ | 1e–8 | 1e–8 |

Table 16: Hyperparameters for Kim et al. (2023)'s Transformer reconstruction model on WikiHan and WikiHan-aug. For the hyperparameters used to train the same model on Hóu, Rom-phon, and Rom-orth, refer to Kim et al. (2023).

| **Dataset** | **Reranking System** | $k$ (beam size) | $\lambda$ (score adjustment weight) |
|---|---|---|---|
| WikiHan | GRU-BS + GRU Reranker | 6 | 1.395 |
|  | GRU-BS + Trans Reranker 1 | 6 | 1.275 |
|  | GRU-BS + Trans Reranker 2 | 7 | 1.260 |
| WikiHan-aug | GRU-BS + GRU Reranker | 7 | 1.620 |
|  | GRU-BS + Trans Reranker 1 | 8 | 1.755 |
|  | GRU-BS + Trans Reranker 2 | 7 | 1.575 |
| Hóu | GRU-BS + GRU Reranker | 7 | 1.755 |
|  | GRU-BS + Trans Reranker 1 | 7 | 2.430 |
|  | GRU-BS + Trans Reranker 2 | 7 | 2.415 |
| Rom-phon | GRU-BS + GRU Reranker | 5 | 0.420 |
|  | GRU-BS + Trans Reranker 1 | 6 | 0.555 |
|  | GRU-BS + Trans Reranker 2 | 6 | 0.585 |
| Rom-orth | GRU-BS + GRU Reranker | 6 | 0.870 |
|  | GRU-BS + Trans Reranker 1 | 6 | 0.990 |
|  | GRU-BS + Trans Reranker 2 | 5 | 0.915 |

Table 17: Reranking hyperparameter search results.

| Dataset | Model | ACC%↑ | TED↓ | TER↓ | FER↓ | BCFS↑ |
|---|---|---|---|---|---|---|
| WikiHan | GRU (baseline) | 66.43% ± 0.29% | 0.5244 ± 0.0049 | 0.1547 ± 0.0014 | 0.0400 ± 0.0006 | 0.7394 ± 0.0022 |
| | Transformer (Kim et al., 2023) | 66.39% ± 0.53% | 0.5302 ± 0.0089 | 0.1564 ± 0.0026 | 0.0406 ± 0.0007 | 0.7370 ± 0.0040 |
| | GRU (Arora et al., 2023) | 64.45% ± 0.34% | 0.5558 ± 0.0060 | 0.1640 ± 0.0018 | 0.0428 ± 0.0007 | 0.7260 ± 0.0027 |
| | Transformer (Arora et al., 2023) | **67.64% ± 0.35%** | **0.5128 ± 0.0072** | **0.1513 ± 0.0021** | **0.0390 ± 0.0006** | **0.7445 ± 0.0031** |
| WikiHan-aug | GRU (baseline) | 68.11% ± 0.44% | 0.5007 ± 0.0083 | 0.1477 ± 0.0024 | 0.0380 ± 0.0007 | 0.7495 ± 0.0036 |
| | Transformer (Kim et al., 2023) | 68.96% ± 0.36% | 0.4889 ± 0.0055 | 0.1442 ± 0.0016 | 0.0371 ± 0.0006 | 0.7551 ± 0.0022 |
| | GRU (Arora et al., 2023) | 66.94% ± 0.68% | 0.5159 ± 0.0101 | 0.1522 ± 0.0030 | 0.0391 ± 0.0011 | 0.7430 ± 0.0043 |
| | Transformer (Arora et al., 2023) | **69.37% ± 0.18%** | **0.4826 ± 0.0028** | **0.1424 ± 0.0008** | **0.0363 ± 0.0004** | **0.7572 ± 0.0013** |
| Hóu | GRU (baseline) | 51.72% ± 0.70% | 0.7777 ± 0.0132 | 0.2037 ± 0.0035 | 0.0488 ± 0.0010 | 0.6783 ± 0.0046 |
| | Transformer (Kim et al., 2023) | 55.46% ± 1.23% | 0.7576 ± 0.0243 | 0.1985 ± 0.0064 | 0.0494 ± 0.0018 | 0.6882 ± 0.0078 |
| | GRU (Arora et al., 2023) | 49.26% ± 1.57% | 0.8266 ± 0.0370 | 0.2166 ± 0.0097 | 0.0528 ± 0.0030 | 0.6622 ± 0.0120 |
| | Transformer (Arora et al., 2023) | **55.60% ± 1.30%** | **0.7520 ± 0.0243** | **0.1970 ± 0.0064** | **0.0485 ± 0.0018** | **0.6892 ± 0.0081** |
| Rom-phon | GRU (baseline) | 63.85% ± 0.37% | 0.7439 ± 0.0068 | 0.1014 ± 0.0009 | **0.0426 ± 0.0005** | 0.8361 ± 0.0014 |
| | Transformer (Kim et al., 2023) | **64.19% ± 0.64%** | **0.7349 ± 0.0096** | **0.1002 ± 0.0013** | 0.0427 ± 0.0006 | **0.8380 ± 0.0019** |
| | GRU (Arora et al., 2023) | 48.28% ± 14.82% | 1.3257 ± 0.8784 | 0.1808 ± 0.1198 | 0.0930 ± 0.0748 | 0.7567 ± 0.1035 |
| | Transformer (Arora et al., 2023) | 63.96% ± 0.65% | 0.7442 ± 0.0087 | 0.1015 ± 0.0012 | 0.0428 ± 0.0005 | 0.8361 ± 0.0018 |
| Rom-orth | GRU (baseline) | 64.58% ± 0.34% | 0.7301 ± 0.0069 | 0.0967 ± 0.0009 | - | 0.8465 ± 0.0014 |
| | Transformer (Kim et al., 2023) | 64.80% ± 0.50% | 0.7258 ± 0.0061 | 0.0961 ± 0.0008 | - | **0.8478 ± 0.0011** |
| | GRU (Arora et al., 2023) | 57.92% ± 2.31% | 0.8741 ± 0.0346 | 0.1158 ± 0.0046 | - | 0.8218 ± 0.0054 |
| | Transformer (Arora et al., 2023) | **65.20% ± 0.46%** | **0.7247 ± 0.0069** | **0.0960 ± 0.0009** | - | 0.8476 ± 0.0012 |

Table 18: Performance means and standard deviations of the reflex prediction models across 20 runs, with the best-performing model for each metric in bold.

| Dataset | Reconstruction System | ACC%↑ | TED↓ | TER↓ | FER↓ | BCFS↑ |
|---|---|---|---|---|---|---|
| WikiHan | GRU (Meloni et al., 2021) | 55.58% ± 0.86% | 0.7360 ± 0.0137 | 0.1724 ± 0.0032 | 0.0686 ± 0.0026 | 0.7426 ± 0.0038 |
| | Trans (Kim et al., 2023) | 54.62% ± 1.22% | 0.7453 ± 0.0165 | 0.1746 ± 0.0039 | 0.0696 ± 0.0029 | 0.7393 ± 0.0048 |
| | GRU-BS ($k = 10$) | 54.88% ± 1.07% | 0.7507 ± 0.0186 | 0.1758 ± 0.0043 | 0.0701 ± 0.0022 | 0.7364 ± 0.0064 |
| | GRU-BS ($k \leq 10$) + GRU Reranker | 57.14% ± 0.80%*† | 0.7045 ± 0.0146*† | 0.1650 ± 0.0034*† | 0.0661 ± 0.0018*† | 0.7515 ± 0.0048*† |
| | GRU-BS ($k \leq 10$) + Trans Reranker 2 | **57.26% ± 0.83%*†** | **0.7029 ± 0.0161*†** | **0.1646 ± 0.0038*†** | **0.0658 ± 0.0021*†** | **0.7520 ± 0.0052*†** |
| WikiHan-aug | GRU (Meloni et al., 2021) | 54.73% ± 0.84% | 0.7574 ± 0.0127 | 0.1774 ± 0.0030 | 0.0689 ± 0.0017 | 0.7346 ± 0.0048 |
| | Trans (Kim et al., 2023) | 55.82% ± 0.97% | 0.7317 ± 0.0165 | 0.1714 ± 0.0039 | 0.0661 ± 0.0020 | 0.7416 ± 0.0053 |
| | GRU-BS ($k = 10$) | 56.64% ± 0.66%* | 0.7214 ± 0.0113 | 0.1690 ± 0.0026 | 0.0658 ± 0.0014 | 0.7454 ± 0.0035 |
| | GRU-BS ($k \leq 10$) + GRU Reranker | **58.58% ± 0.70%*†** | **0.6822 ± 0.0143*†** | **0.1598 ± 0.0033*†** | 0.0628 ± 0.0017*† | **0.7579 ± 0.0040*†** |
| | GRU-BS ($k \leq 10$) + Trans Reranker 2 | 58.58% ± 0.75%*† | 0.6840 ± 0.0129*† | 0.1602 ± 0.0030*† | **0.0626 ± 0.0017*†** | 0.7575 ± 0.0038*† |
| Hóu | GRU (Meloni et al., 2021) | 34.63% ± 2.37% | 1.0916 ± 0.0629 | 0.2479 ± 0.0147 | 0.0914 ± 0.0049 | 0.6697 ± 0.0167 |
| | Trans (Kim et al., 2023) | 39.01% ± 2.89% | 0.9904 ± 0.0443 | 0.2233 ± 0.0108 | 0.0875 ± 0.0069 | 0.6955 ± 0.0103 |
| | GRU-BS ($k = 10$) | 37.36% ± 3.25% | 1.0382 ± 0.0662 | 0.2328 ± 0.0148 | 0.0917 ± 0.0065 | 0.6974 ± 0.0176 |
| | GRU-BS ($k \leq 10$) + GRU Reranker | 40.50% ± 3.09%† | 0.9727 ± 0.0486† | 0.2181 ± 0.0109† | 0.0867 ± 0.0058† | 0.7130 ± 0.0132*† |
| | GRU-BS ($k \leq 10$) + Trans Reranker 2 | **42.08% ± 2.96%*†** | **0.9503 ± 0.0525*†** | **0.2131 ± 0.0118*†** | **0.0850 ± 0.0063†** | **0.7170 ± 0.0137*†** |
| Rom-phon | GRU (Meloni et al., 2021) | 51.92% ± 0.65% | 0.9775 ± 0.0216 | 0.1244 ± 0.0028 | 0.0390 ± 0.0012 | 0.8275 ± 0.0033 |
| | Trans (Kim et al., 2023) | 53.04% ± 0.80% | 0.9050 ± 0.0166 | 0.1148 ± 0.0018 | 0.0377 ± 0.0008 | 0.8417 ± 0.0024 |
| | GRU-BS ($k = 10$) | 52.63% ± 0.68% | 0.9125 ± 0.0174 | 0.1018 ± 0.0019* | 0.0353 ± 0.0009* | 0.8402 ± 0.0032 |
| | GRU-BS ($k \leq 10$) + GRU Reranker | **53.95% ± 0.77%*†** | 0.8775 ± 0.0165*† | 0.0979 ± 0.0018*† | 0.0336 ± 0.0007*† | 0.8460 ± 0.0028*† |
| | GRU-BS ($k \leq 10$) + Trans Reranker 2 | 53.85% ± 0.79%*† | **0.8765 ± 0.0177*†** | **0.0978 ± 0.0020*†** | **0.0333 ± 0.0008*†** | **0.8461 ± 0.0030*†** |
| Rom-orth | GRU (Meloni et al., 2021) | 69.41% ± 0.53% | 0.6004 ± 0.0130 | 0.0781 ± 0.0018 | - | 0.8906 ± 0.0023 |
| | Trans (Kim et al., 2023) | 71.05% ± 0.50% | 0.5636 ± 0.0163 | 0.0734 ± 0.0022 | - | 0.8981 ± 0.0028 |
| | GRU-BS ($k = 10$) | 71.09% ± 0.51% | 0.5531 ± 0.0127 | 0.0617 ± 0.0014* | - | 0.8990 ± 0.0023 |
| | GRU-BS ($k \leq 10$) + GRU Reranker | **72.60% ± 0.41%*†** | **0.5237 ± 0.0109*†** | **0.0584 ± 0.0012*†** | - | **0.9045 ± 0.0019*†** |
| | GRU-BS ($k \leq 10$) + Trans Reranker 2 | 72.50% ± 0.45%*† | 0.5246 ± 0.0111*† | 0.0585 ± 0.0012*† | - | 0.9044 ± 0.0020*† |

Table 19: Performance means and standard deviations of reconstruction systems across 20 runs. Reconstruction systems include baselines, GRU with beam search (GRU-BS), and GRU-BS with reranking. Bold indicates the best-performing system for each metric, asterisks indicate statistically better performance than both baseline models (Meloni et al. (2021)'s GRU and Kim et al. (2023)'s Transformer), and daggers indicate that a reranking system performs statistically better than its beam search counterpart.

|           | Cantonese | Gan   | Hakka | Jin   | Mandarin | Hokkien | Wu    | Xiang |
|-----------|-----------|-------|-------|-------|----------|---------|-------|-------|
| Not-in    | 75.3%     | 72.7% | 70.8% | 83.3% | 72.2%    | 82.1%   | 57.7% | 82.1% |
| Worsened  | 59.4%     | 75.0% | 40.0% | 71.4% | 68.8%    | 68.8%   | 33.3% | 60.0% |
| Unchanged | 16.6%     | 19.1% | 32.2% | 14.5% | 15.1%    | 39.5%   | 14.1% | 29.7% |
| Improved  | 34.9%     | 30.0% | 51.0% | 10.0% | 21.4%    | 48.7%   | 31.6% | 38.1% |
| Overall   | 28.6%     | 26.5% | 38.8% | 23.7% | 26.2%    | 47.8%   | 20.4% | 37.3% |

Table 20: Transformer reranker error rates (when predicting reflexes from the gold protoform) for each daughter language in the WikiHan dataset given each reranker behavior category, obtained from a randomly selected run.

| Model 1 | Model 2 | ACC%↑ | TED↓ | TER↓ | FER↓ | BCFS↑ |
|---|---|---|---|---|---|---|
| GRU (baseline) | Trans (Kim et al., 2023) | $p = 0.4196$ (−0.0029, 0.0038) | $p = 0.0093^*$ (−0.0112, 0.0001) | $p = 0.0093^*$ (−0.0033, 0.0000) | $p = 0.0021^*$ (−0.0011, −0.0001)* | $p = 0.0242$ (−0.0003, 0.0047) |
| | GRU (Arora et al., 2023) | $p < 0.0001^*$ (0.0173, 0.0225)* | $p < 0.0001^*$ (−0.0355, −0.0268)* | $p < 0.0001^*$ (−0.0105, −0.0079)* | $p < 0.0001^*$ (−0.0033, −0.0022)* | $p < 0.0001^*$ (0.0113, 0.0152)* |
| | Trans (Arora et al., 2023) | $p = 1.0000$ (−0.0149, −0.0095) | $p = 1.0000$ (0.0060, 0.0166) | $p = 1.0000$ (0.0018, 0.0049) | $p = 1.0000$ (0.0005, 0.0015) | $p = 1.0000$ (−0.0073, −0.0027) |
| Trans (Kim et al., 2023) | GRU (baseline) | $p = 0.5804$ (−0.0040, 0.0031) | $p = 0.9907$ (−0.0003, 0.0115) | $p = 0.9907$ (−0.0001, 0.0034) | $p = 0.9979$ (0.0001, 0.0011) | $p = 0.9758$ (−0.0049, 0.0004) |
| | GRU (Arora et al., 2023) | $p < 0.0001^*$ (0.0158, 0.0230)* | $p < 0.0001^*$ (−0.0318, −0.0195)* | $p < 0.0001^*$ (−0.0094, −0.0058)* | $p < 0.0001^*$ (−0.0027, −0.0016)* | $p < 0.0001^*$ (0.0083, 0.0138)* |
| | Trans (Arora et al., 2023) | $p = 1.0000$ (−0.0162, −0.0089) | $p = 1.0000$ (0.0105, 0.0236) | $p = 1.0000$ (0.0031, 0.0070) | $p = 1.0000$ (0.0011, 0.0021) | $p = 1.0000$ (−0.0102, −0.0044) |
| GRU (Arora et al., 2023) | GRU (baseline) | $p = 1.0000$ (−0.0223, −0.0173) | $p = 1.0000$ (0.0265, 0.0355) | $p = 1.0000$ (0.0078, 0.0105) | $p = 1.0000$ (0.0022, 0.0033) | $p = 1.0000$ (−0.0152, −0.0112) |
| | Trans (Kim et al., 2023) | $p = 1.0000$ (−0.0229, −0.0158) | $p = 1.0000$ (0.0196, 0.0317) | $p = 1.0000$ (0.0058, 0.0093) | $p = 1.0000$ (0.0016, 0.0027) | $p = 1.0000$ (−0.0138, −0.0084) |
| | Trans (Arora et al., 2023) | $p = 1.0000$ (−0.0348, −0.0291) | $p = 1.0000$ (0.0372, 0.0482) | $p = 1.0000$ (0.0110, 0.0142) | $p = 1.0000$ (0.0032, 0.0043) | $p = 1.0000$ (−0.0208, −0.0160) |
| Trans (Arora et al., 2023) | GRU (baseline) | $p < 0.0001^*$ (0.0096, 0.0148)* | $p < 0.0001^*$ (−0.0165, −0.0063)* | $p < 0.0001^*$ (−0.0049, −0.0019)* | $p < 0.0001^*$ (−0.0015, −0.0005)* | $p < 0.0001^*$ (0.0029, 0.0073)* |
| | Trans (Kim et al., 2023) | $p < 0.0001^*$ (0.0090, 0.0161)* | $p < 0.0001^*$ (−0.0236, −0.0105)* | $p < 0.0001^*$ (−0.0070, −0.0031)* | $p < 0.0001^*$ (−0.0021, −0.0011)* | $p < 0.0001^*$ (0.0044, 0.0102)* |
| | GRU (Arora et al., 2023) | $p < 0.0001^*$ (0.0293, 0.0349)* | $p < 0.0001^*$ (−0.0482, −0.0371)* | $p < 0.0001^*$ (−0.0142, −0.0110)* | $p < 0.0001^*$ (−0.0043, −0.0032)* | $p < 0.0001^*$ (0.0159, 0.0208)* |

Table 21: Reflex prediction significance test results for WikiHan. Asterisks indicates that Model 1 performs better than Model 2 with the corresponding test (*p*-value or CI).

| Model 1 | Model 2 | ACC%↑ | TED↓ | TER↓ | FER↓ | BCFS↑ |
|---|---|---|---|---|---|---|
| GRU (baseline) | Trans (Kim et al., 2023) | $p = 1.0000$ (−0.0120, −0.0053) | $p = 1.0000$ (0.0066, 0.0182) | $p = 1.0000$ (0.0020, 0.0054) | $p = 1.0000$ (0.0005, 0.0015) | $p = 1.0000$ (−0.0083, −0.0034) |
| | GRU (Arora et al., 2023) | $p < 0.0001^*$ (0.0074, 0.0166)* | $p < 0.0001^*$ (−0.0224, −0.0075)* | $p < 0.0001^*$ (−0.0066, −0.0022)* | $p < 0.0010^*$ (−0.0018, −0.0003)* | $p < 0.0001^*$ (0.0033, 0.0096)* |
| | Trans (Arora et al., 2023) | $p = 1.0000$ (−0.0158, −0.0101) | $p = 1.0000$ (0.0139, 0.0245) | $p = 1.0000$ (0.0041, 0.0072) | $p = 1.0000$ (0.0013, 0.0022) | $p = 1.0000$ (−0.0104, −0.0058) |
| Trans (Kim et al., 2023) | GRU (baseline) | $p < 0.0001^*$ (0.0053, 0.0121)* | $p < 0.0001^*$ (−0.0183, −0.0066)* | $p < 0.0001^*$ (−0.0054, −0.0019)* | $p < 0.0001^*$ (−0.0015, −0.0005)* | $p < 0.0001^*$ (0.0034, 0.0083)* |
| | GRU (Arora et al., 2023) | $p < 0.0001^*$ (0.0162, 0.0252)* | $p < 0.0001^*$ (−0.0342, −0.0207)* | $p < 0.0001^*$ (−0.0101, −0.0061)* | $p < 0.0001^*$ (−0.0028, −0.0013)* | $p < 0.0001^*$ (0.0094, 0.0151)* |
| | Trans (Arora et al., 2023) | $p = 0.9999$ (−0.0067, −0.0020) | $p = 1.0000$ (0.0030, 0.0102) | $p = 1.0000$ (0.0009, 0.0030) | $p = 1.0000$ (0.0004, 0.0012) | $p = 0.9994$ (−0.0037, −0.0007) |
| GRU (Arora et al., 2023) | GRU (baseline) | $p = 1.0000$ (−0.0169, −0.0073) | $p = 1.0000$ (0.0075, 0.0230) | $p = 1.0000$ (0.0022, 0.0068) | $p = 0.9997$ (0.0003, 0.0019) | $p = 1.0000$ (−0.0098, −0.0032) |
| | Trans (Kim et al., 2023) | $p = 1.0000$ (−0.0250, −0.0162) | $p = 1.0000$ (0.0208, 0.0339) | $p = 1.0000$ (0.0061, 0.0100) | $p = 1.0000$ (0.0013, 0.0028) | $p = 1.0000$ (−0.0150, −0.0095) |
| | Trans (Arora et al., 2023) | $p = 1.0000$ (−0.0290, −0.0209) | $p = 1.0000$ (0.0279, 0.0400) | $p = 1.0000$ (0.0082, 0.0118) | $p = 1.0000$ (0.0022, 0.0035) | $p = 1.0000$ (−0.0170, −0.0118) |
| Trans (Arora et al., 2023) | GRU (baseline) | $p < 0.0001^*$ (0.0102, 0.0158)* | $p < 0.0001^*$ (−0.0244, −0.0140)* | $p < 0.0001^*$ (−0.0072, −0.0041)* | $p < 0.0001^*$ (−0.0022, −0.0013)* | $p < 0.0001^*$ (0.0058, 0.0104)* |
| | Trans (Kim et al., 2023) | $p < 0.0001^*$ (0.0019, 0.0065)* | $p < 0.0001^*$ (−0.0102, −0.0029)* | $p < 0.0001^*$ (−0.0030, −0.0009)* | $p < 0.0001^*$ (−0.0012, −0.0004)* | $p < 0.0010^*$ (0.0007, 0.0037)* |
| | GRU (Arora et al., 2023) | $p < 0.0001^*$ (0.0208, 0.0290)* | $p < 0.0001^*$ (−0.0398, −0.0277)* | $p < 0.0001^*$ (−0.0117, −0.0082)* | $p < 0.0001^*$ (−0.0035, −0.0022)* | $p < 0.0001^*$ (0.0118, 0.0169)* |

Table 22: Reflex prediction significance test results for WikiHan-aug. Asterisks indicates that Model 1 performs better than Model 2 with the corresponding test (*p*-value or CI).

| Model 1 | Model 2 | ACC%↑ | TED↓ | TER↓ | FER↓ | BCFS↑ |
|---|---|---|---|---|---|---|
| GRU (baseline) | Trans (Kim et al., 2023) | $p = 1.0000$ (−0.0453, −0.0292) | $p = 0.9989$ (0.0037, 0.0360) | $p = 0.9989$ (0.0010, 0.0094) | $p = 0.1337$ (−0.0017, 0.0006) | $p = 1.0000$ (−0.0149, −0.0044) |
| | GRU (Arora et al., 2023) | $p < 0.0001^*$ (0.0149, 0.0342)* | $p < 0.0001^*$ (−0.0709, −0.0276)* | $p < 0.0001^*$ (−0.0186, −0.0072)* | $p < 0.0001^*$ (−0.0057, −0.0022)* | $p < 0.0001^*$ (0.0091, 0.0235)* |
| | Trans (Arora et al., 2023) | $p = 1.0000$ (−0.0474, −0.0305) | $p = 0.9998$ (0.0097, 0.0414) | $p = 0.9998$ (0.0025, 0.0109) | $p = 0.9075$ (−0.0009, 0.0014) | $p = 1.0000$ (−0.0161, −0.0053) |
| Trans (Kim et al., 2023) | GRU (baseline) | $p < 0.0001^*$ (0.0294, 0.0455)* | $p = 0.0011^*$ (−0.0355, −0.0042)* | $p = 0.0011^*$ (−0.0093, −0.0011)* | $p = 0.8663$ (−0.0006, 0.0017) | $p < 0.0001^*$ (0.0046, 0.0150)* |
| | GRU (Arora et al., 2023) | $p < 0.0001^*$ (0.0507, 0.0732)* | $p < 0.0001^*$ (−0.0949, −0.0441)* | $p < 0.0001^*$ (−0.0249, −0.0116)* | $p < 0.0010^*$ (−0.0054, −0.0015)* | $p < 0.0001^*$ (0.0180, 0.0343)* |
| | Trans (Arora et al., 2023) | $p = 0.6221$ (−0.0118, 0.0086) | $p = 0.8066$ (−0.0137, 0.0254) | $p = 0.8066$ (−0.0036, 0.0067) | $p = 0.9384$ (−0.0006, 0.0023) | $p = 0.6964$ (−0.0076, 0.0053) |
| GRU (Arora et al., 2023) | GRU (baseline) | $p = 1.0000$ (−0.0340, −0.0147) | $p = 1.0000$ (0.0272, 0.0711) | $p = 1.0000$ (0.0071, 0.0186) | $p = 1.0000$ (0.0022, 0.0058) | $p = 1.0000$ (−0.0234, −0.0090) |
| | Trans (Kim et al., 2023) | $p = 1.0000$ (−0.0733, −0.0505) | $p = 1.0000$ (0.0444, 0.0942) | $p = 1.0000$ (0.0116, 0.0247) | $p = 0.9998$ (0.0014, 0.0054) | $p = 1.0000$ (−0.0341, −0.0181) |
| | Trans (Arora et al., 2023) | $p = 1.0000$ (−0.0750, −0.0518) | $p = 1.0000$ (0.0495, 0.0999) | $p = 1.0000$ (0.0130, 0.0262) | $p = 1.0000$ (0.0023, 0.0063) | $p = 1.0000$ (−0.0353, −0.0187) |
| Trans (Arora et al., 2023) | GRU (baseline) | $p < 0.0001^*$ (0.0301, 0.0472)* | $p < 0.0010^*$ (−0.0416, −0.0098)* | $p < 0.0010^*$ (−0.0109, −0.0026)* | $p = 0.0925$ (−0.0014, 0.0010) | $p < 0.0001^*$ (0.0054, 0.0162)* |
| | Trans (Kim et al., 2023) | $p = 0.3779$ (−0.0087, 0.0120) | $p = 0.1934$ (−0.0256, 0.0147) | $p = 0.1934$ (−0.0067, 0.0039) | $p = 0.0616$ (−0.0023, 0.0007) | $p = 0.3036$ (−0.0056, 0.0076) |
| | GRU (Arora et al., 2023) | $p < 0.0001^*$ (0.0518, 0.0749)* | $p < 0.0001^*$ (−0.0995, −0.0494)* | $p < 0.0001^*$ (−0.0261, −0.0129)* | $p < 0.0001^*$ (−0.0062, −0.0024)* | $p < 0.0001^*$ (0.0188, 0.0351)* |

Table 23: Reflex prediction significance test results for Hóu. Asterisks indicates that Model 1 performs better than Model 2 with the corresponding test (*p*-value or CI).

| Model 1 | Model 2 | ACC%↑ | TED↓ | TER↓ | FER↓ | BCFS↑ |
|---|---|---|---|---|---|---|
| GRU (baseline) | Trans (Kim et al., 2023) | $p = 0.9848$ (−0.0076, 0.0010) | $p = 0.9983$ (0.0022, 0.0158) | $p = 0.9983$ (0.0003, 0.0022) | $p = 0.4143$ (−0.0005, 0.0004) | $p = 0.9983$ (−0.0032, −0.0005) |
| | GRU (Arora et al., 2023) | $p < 0.0001$* (0.0956, 0.2815)* | $p < 0.0001$* (−1.3739, −0.2318)* | $p < 0.0001$* (−0.1874, −0.0316)* | $p < 0.0001$* (−0.1226, −0.0206)* | $p < 0.0001$* (0.0392, 0.1708)* |
| | Trans (Arora et al., 2023) | $p = 0.8066$ (−0.0051, 0.0034) | $p = 0.3132$ (−0.0064, 0.0064) | $p = 0.3132$ (−0.0009, 0.0009) | $p = 0.0616$ (−0.0005, 0.0002) | $p = 0.3727$ (−0.0014, 0.0012) |
| Trans (Kim et al., 2023) | GRU (baseline) | $p = 0.0152$ (−0.0010, 0.0075) | $p = 0.0017$* (−0.0160, −0.0022)* | $p = 0.0017$* (−0.0022, −0.0003)* | $p = 0.5857$ (−0.0004, 0.0005) | $p = 0.0017$* (0.0005, 0.0033)* |
| | GRU (Arora et al., 2023) | $p < 0.0001$* (0.0989, 0.2843)* | $p < 0.0001$* (−1.3808, −0.2415)* | $p < 0.0001$* (−0.1883, −0.0329)* | $p < 0.0001$* (−0.1225, −0.0206)* | $p < 0.0001$* (0.0411, 0.1726)* |
| | Trans (Arora et al., 2023) | $p = 0.2085$ (−0.0030, 0.0076) | $p = 0.0027$* (−0.0167, −0.0017)* | $p = 0.0027$* (−0.0023, −0.0002)* | $p = 0.1789$ (−0.0006, 0.0003) | $p = 0.0043$* (0.0003, 0.0033)* |
| GRU (Arora et al., 2023) | GRU (baseline) | $p = 1.0000$ (−0.2892, −0.0959) | $p = 1.0000$ (0.2447, 1.3977) | $p = 1.0000$ (0.0334, 0.1906) | $p = 1.0000$ (0.0214, 0.1162) | $p = 1.0000$ (−0.1743, −0.0405) |
| | Trans (Kim et al., 2023) | $p = 1.0000$ (−0.2924, −0.0998) | $p = 1.0000$ (0.2544, 1.4058) | $p = 1.0000$ (0.0347, 0.1917) | $p = 1.0000$ (0.0213, 0.1160) | $p = 1.0000$ (−0.1760, −0.0423) |
| | Trans (Arora et al., 2023) | $p = 1.0000$ (−0.2885, −0.0971) | $p = 1.0000$ (0.2443, 1.3977) | $p = 1.0000$ (0.0333, 0.1906) | $p = 1.0000$ (0.0211, 0.1160) | $p = 1.0000$ (−0.1744, −0.0404) |
| Trans (Arora et al., 2023) | GRU (baseline) | $p = 0.1934$ (−0.0035, 0.0053) | $p = 0.6868$ (−0.0064, 0.0065) | $p = 0.6868$ (−0.0009, 0.0009) | $p = 0.9384$ (−0.0002, 0.0005) | $p = 0.6273$ (−0.0013, 0.0014) |
| | Trans (Kim et al., 2023) | $p = 0.7915$ (−0.0077, 0.0031) | $p = 0.9973$ (0.0015, 0.0167) | $p = 0.9973$ (0.0002, 0.0023) | $p = 0.8211$ (−0.0003, 0.0006) | $p = 0.9957$ (−0.0033, −0.0002) |
| | GRU (Arora et al., 2023) | $p < 0.0001$* (0.0963, 0.2810)* | $p < 0.0001$* (−1.3713, −0.2328)* | $p < 0.0001$* (−0.1870, −0.0317)* | $p < 0.0001$* (−0.1225, −0.0204)* | $p < 0.0001$* (0.0392, 0.1710)* |

Table 24: Reflex prediction significance test results for Rom-phon. Asterisks indicates that Model 1 performs better than Model 2 with the corresponding test (*p*-value or CI).

| Model 1 | Model 2 | ACC%↑ | TED↓ | TER↓ | FER↓ | BCFS↑ |
|---|---|---|---|---|---|---|
| GRU (baseline) | Trans (Kim et al., 2023) | $p = 0.9837$ (−0.0052, 0.0020) | $p = 0.9661$ (−0.0012, 0.0094) | $p = 0.9661$ (−0.0002, 0.0012) | - | $p = 0.9971$ (−0.0022, −0.0003) |
| | GRU (Arora et al., 2023) | $p < 0.0001$* (0.0559, 0.0829)* | $p < 0.0001$* (−0.1649, −0.1257)* | $p < 0.0001$* (−0.0218, −0.0166)* | - | $p < 0.0001$* (0.0218, 0.0281)* |
| | Trans (Arora et al., 2023) | $p = 1.0000$ (−0.0095, −0.0028) | $p = 0.9848$ (−0.0003, 0.0110) | $p = 0.9848$ (−0.0000, 0.0015) | - | $p = 0.9893$ (−0.0022, −0.0000) |
| Trans (Kim et al., 2023) | GRU (baseline) | $p = 0.0163$ (−0.0021, 0.0052) | $p = 0.0339$ (−0.0092, 0.0012) | $p = 0.0339$ (−0.0012, 0.0002) | - | $p = 0.0029$* (0.0003, 0.0022)* |
| | GRU (Arora et al., 2023) | $p < 0.0001$* (0.0577, 0.0850)* | $p < 0.0001$* (−0.1690, −0.1299)* | $p < 0.0001$* (−0.0224, −0.0172)* | - | $p < 0.0001$* (0.0230, 0.0292)* |
| | Trans (Arora et al., 2023) | $p = 0.9951$ (−0.0083, −0.0005) | $p = 0.6118$ (−0.0039, 0.0068) | $p = 0.6118$ (−0.0005, 0.0009) | - | $p = 0.3132$ (−0.0008, 0.0011) |
| GRU (Arora et al., 2023) | GRU (baseline) | $p = 1.0000$ (−0.0826, −0.0555) | $p = 1.0000$ (0.1246, 0.1651) | $p = 1.0000$ (0.0165, 0.0219) | - | $p = 1.0000$ (−0.0280, −0.0216) |
| | Trans (Kim et al., 2023) | $p = 1.0000$ (−0.0850, −0.0576) | $p = 1.0000$ (0.1294, 0.1689) | $p = 1.0000$ (0.0171, 0.0224) | - | $p = 1.0000$ (−0.0292, −0.0230) |
| | Trans (Arora et al., 2023) | $p = 1.0000$ (−0.0890, −0.0617) | $p = 1.0000$ (0.1301, 0.1702) | $p = 1.0000$ (0.0172, 0.0225) | - | $p = 1.0000$ (−0.0291, −0.0228) |
| Trans (Arora et al., 2023) | GRU (baseline) | $p < 0.0001$* (0.0027, 0.0094)* | $p = 0.0152$ (−0.0109, 0.0002) | $p = 0.0152$ (−0.0014, 0.0000) | - | $p = 0.0107$ (0.0001, 0.0022)* |
| | Trans (Kim et al., 2023) | $p = 0.0049$* (0.0003, 0.0085)* | $p = 0.3882$ (−0.0067, 0.0041) | $p = 0.3882$ (−0.0009, 0.0005) | - | $p = 0.6868$ (−0.0011, 0.0008) |
| | GRU (Arora et al., 2023) | $p < 0.0001$* (0.0619, 0.0894)* | $p < 0.0001$* (−0.1702, −0.1308)* | $p < 0.0001$* (−0.0225, −0.0173)* | - | $p < 0.0001$* (0.0229, 0.0291)* |

Table 25: Reflex prediction significance test results for Rom-orth. Asterisks indicates that Model 1 performs better than Model 2 with the corresponding test (*p*-value or CI).

| Reconstruction System 1 | Reconstruction System 2 | ACC%↑ | TED↓ | TER↓ | FER↓ | BCFS↑ |
|---|---|---|---|---|---|---|
| GRU (Meloni et al., 2021) | Trans (Kim et al., 2023) | $p = 0.0075^*$ (0.0011, 0.0185)* | $p = 0.0283$ (−0.0217, 0.0029) | $p = 0.0283$ (−0.0051, 0.0007) | $p = 0.1719$ (−0.0034, 0.0011) | $p = 0.0080^*$ (−0.0004, 0.0067) |
| | GRU-BS | $p = 0.0405$ (−0.0009, 0.0152) | $p = 0.0040^*$ (−0.0288, −0.0018)* | $p = 0.0040^*$ (−0.0068, −0.0004)* | $p = 0.0442$ (−0.0035, 0.0004) | $p < 0.0010^*$ (0.0023, 0.0111)* |
| | GRU-BS + GRU Reranker | $p = 1.0000$ (−0.0223, −0.0090) | $p = 1.0000$ (0.0196, 0.0428) | $p = 1.0000$ (0.0046, 0.0100) | $p = 0.9992$ (0.0006, 0.0042) | $p = 1.0000$ (−0.0122, −0.0050) |
| | GRU-BS + Trans. Reranker | $p = 1.0000$ (−0.0240, −0.0101) | $p = 1.0000$ (0.0216, 0.0464) | $p = 1.0000$ (0.0051, 0.0109) | $p = 0.9991$ (0.0009, 0.0047) | $p = 1.0000$ (−0.0134, −0.0057) |
| Trans (Kim et al., 2023) | GRU (Meloni et al., 2021) | $p = 0.9925$ (−0.0182, −0.0014) | $p = 0.9717$ (−0.0023, 0.0222) | $p = 0.9717$ (−0.0005, 0.0052) | $p = 0.8281$ (−0.0011, 0.0034) | $p = 0.9920$ (−0.0067, 0.0002) |
| | GRU-BS | $p = 0.8066$ (−0.0124, 0.0063) | $p = 0.1427$ (−0.0197, 0.0092) | $p = 0.1427$ (−0.0046, 0.0022) | $p = 0.2581$ (−0.0025, 0.0016) | $p = 0.0684$ (−0.0015, 0.0077) |
| | GRU-BS + GRU Reranker | $p = 1.0000$ (−0.0339, −0.0173) | $p = 1.0000$ (0.0280, 0.0533) | $p = 1.0000$ (0.0066, 0.0125) | $p = 0.9999$ (0.0017, 0.0056) | $p = 1.0000$ (−0.0158, −0.0080) |
| | GRU-BS + Trans. Reranker | $p = 1.0000$ (−0.0354, −0.0183) | $p = 1.0000$ (0.0302, 0.0570) | $p = 1.0000$ (0.0071, 0.0134) | $p = 1.0000$ (0.0019, 0.0060) | $p = 1.0000$ (−0.0169, −0.0087) |
| GRU-BS | GRU (Meloni et al., 2021) | $p = 0.9595$ (−0.0150, 0.0007) | $p = 0.9960$ (0.0016, 0.0286) | $p = 0.9960$ (0.0004, 0.0067) | $p = 0.9558$ (−0.0004, 0.0035) | $p = 0.9996$ (−0.0109, −0.0022) |
| | Trans (Kim et al., 2023) | $p = 0.1934$ (−0.0068, 0.0122) | $p = 0.8573$ (−0.0085, 0.0197) | $p = 0.8573$ (−0.0020, 0.0046) | $p = 0.7419$ (−0.0016, 0.0025) | $p = 0.9316$ (−0.0080, 0.0014) |
| | GRU-BS + GRU Reranker | $p = 1.0000$ (−0.0304, −0.0150) | $p = 1.0000$ (0.0327, 0.0607) | $p = 1.0000$ (0.0077, 0.0142) | $p = 1.0000$ (0.0024, 0.0056) | $p = 1.0000$ (−0.0200, −0.0106) |
| | GRU-BS + Trans. Reranker | $p = 1.0000$ (−0.0317, −0.0162) | $p = 1.0000$ (0.0344, 0.0626) | $p = 1.0000$ (0.0081, 0.0147) | $p = 1.0000$ (0.0026, 0.0061) | $p = 1.0000$ (−0.0208, −0.0111) |
| GRU-BS + GRU Reranker | GRU (Meloni et al., 2021) | $p < 0.0001^*$ (0.0090, 0.0224)* | $p < 0.0001^*$ (−0.0426, −0.0193)* | $p < 0.0001^*$ (−0.0100, −0.0045)* | $p < 0.0010^*$ (−0.0042, −0.0006)* | $p < 0.0001^*$ (0.0051, 0.0121)* |
| | Trans (Kim et al., 2023) | $p < 0.0001^*$ (0.0172, 0.0339)* | $p < 0.0001^*$ (−0.0537, −0.0281)* | $p < 0.0001^*$ (−0.0126, −0.0066)* | $p < 0.0001^*$ (−0.0056, −0.0018)* | $p < 0.0001^*$ (0.0078, 0.0158)* |
| | GRU-BS | $p < 0.0001^*$ (0.0151, 0.0303)* | $p < 0.0001^*$ (−0.0598, −0.0321)* | $p < 0.0001^*$ (−0.0140, −0.0075)* | $p < 0.0001^*$ (−0.0056, −0.0024)* | $p < 0.0001^*$ (0.0105, 0.0197)* |
| | GRU-BS + Trans. Reranker | $p = 0.6723$ (−0.0080, 0.0053) | $p = 0.5962$ (−0.0102, 0.0158) | $p = 0.5962$ (−0.0024, 0.0037) | $p = 0.6425$ (−0.0012, 0.0020) | $p = 0.5216$ (−0.0051, 0.0032) |
| GRU-BS + Trans. Reranker | GRU (Meloni et al., 2021) | $p < 0.0001^*$ (0.0099, 0.0237)* | $p < 0.0001^*$ (−0.0456, −0.0213)* | $p < 0.0001^*$ (−0.0107, −0.0050)* | $p < 0.0010^*$ (−0.0047, −0.0009)* | $p < 0.0001^*$ (0.0058, 0.0132)* |
| | Trans (Kim et al., 2023) | $p < 0.0001^*$ (0.0183, 0.0356)* | $p < 0.0001^*$ (−0.0564, −0.0297)* | $p < 0.0001^*$ (−0.0132, −0.0070)* | $p < 0.0001^*$ (−0.0060, −0.0020)* | $p < 0.0001^*$ (0.0085, 0.0169)* |
| | GRU-BS | $p < 0.0001^*$ (0.0160, 0.0314)* | $p < 0.0001^*$ (−0.0625, −0.0338)* | $p < 0.0001^*$ (−0.0146, −0.0079)* | $p < 0.0001^*$ (−0.0061, −0.0026)* | $p < 0.0001^*$ (0.0110, 0.0207)* |
| | GRU-BS + GRU Reranker | $p = 0.3277$ (−0.0054, 0.0080) | $p = 0.4038$ (−0.0150, 0.0100) | $p = 0.4038$ (−0.0035, 0.0023) | $p = 0.3575$ (−0.0019, 0.0012) | $p = 0.4784$ (−0.0034, 0.0050) |

Table 26: Reconstruction significance test results for WikiHan. Asterisks indicates that Reconstruction System 1 performs better than Reconstruction System 2 with the corresponding test (*p*-value or CI).

| Reconstruction System 1 | Reconstruction System 2 | ACC%↑ | TED↓ | TER↓ | FER↓ | BCFS↑ |
|---|---|---|---|---|---|---|
| GRU (Meloni et al., 2021) | Trans (Kim et al., 2023) | $p = 0.9998$ (−0.0171, −0.0025) | $p = 1.0000$ (0.0120, 0.0362) | $p = 1.0000$ (0.0028, 0.0085) | $p = 1.0000$ (0.0012, 0.0042) | $p = 0.9998$ (−0.0108, −0.0025) |
| | GRU-BS | $p = 1.0000$ (−0.0245, −0.0121) | $p = 1.0000$ (0.0253, 0.0449) | $p = 1.0000$ (0.0059, 0.0105) | $p = 1.0000$ (0.0018, 0.0044) | $p = 1.0000$ (−0.0141, −0.0074) |
| | GRU-BS + GRU Reranker | $p = 1.0000$ (−0.0442, −0.0315) | $p = 1.0000$ (0.0633, 0.0854) | $p = 1.0000$ (0.0148, 0.0200) | $p = 1.0000$ (0.0047, 0.0074) | $p = 1.0000$ (−0.0268, −0.0196) |
| | GRU-BS + Trans. Reranker | $p = 1.0000$ (−0.0444, −0.0313) | $p = 1.0000$ (0.0623, 0.0829) | $p = 1.0000$ (0.0146, 0.0194) | $p = 1.0000$ (0.0050, 0.0077) | $p = 1.0000$ (−0.0263, −0.0194) |
| Trans (Kim et al., 2023) | GRU (Meloni et al., 2021) | $p < 0.0010^*$ (0.0022, 0.0171)* | $p < 0.0001^*$ (−0.0361, −0.0125)* | $p < 0.0001^*$ (−0.0084, −0.0029)* | $p < 0.0001^*$ (−0.0042, −0.0012)* | $p < 0.0001^*$ (0.0025, 0.0107)* |
| | GRU-BS | $p = 0.9971$ (−0.0160, −0.0023) | $p = 0.9595$ (0.0001, 0.0226) | $p = 0.9595$ (0.0000, 0.0053) | $p = 0.6723$ (−0.0010, 0.0019) | $p = 0.9801$ (−0.0079, −0.0005) |
| | GRU-BS + GRU Reranker | $p = 1.0000$ (−0.0358, −0.0216) | $p = 1.0000$ (0.0375, 0.0629) | $p = 1.0000$ (0.0088, 0.0147) | $p = 1.0000$ (0.0018, 0.0049) | $p = 1.0000$ (−0.0205, −0.0127) |
| | GRU-BS + Trans. Reranker | $p = 1.0000$ (−0.0360, −0.0212) | $p = 1.0000$ (0.0367, 0.0607) | $p = 1.0000$ (0.0086, 0.0142) | $p = 1.0000$ (0.0021, 0.0052) | $p = 1.0000$ (−0.0200, −0.0126) |
| GRU-BS | GRU (Meloni et al., 2021) | $p < 0.0001^*$ (0.0118, 0.0247)* | $p < 0.0001^*$ (−0.0450, −0.0251)* | $p < 0.0001^*$ (−0.0105, −0.0059)* | $p < 0.0001^*$ (−0.0043, −0.0019)* | $p < 0.0001^*$ (0.0073, 0.0142)* |
| | Trans (Kim et al., 2023) | $p = 0.0029^*$ (0.0022, 0.0158)* | $p = 0.0405$ (−0.0230, −0.0001)* | $p = 0.0405$ (−0.0054, −0.0000)* | $p = 0.3277$ (−0.0019, 0.0010) | $p = 0.0199$ (0.0005, 0.0079)* |
| | GRU-BS + GRU Reranker | $p = 1.0000$ (−0.0251, −0.0141) | $p = 1.0000$ (0.0285, 0.0496) | $p = 1.0000$ (0.0067, 0.0116) | $p = 1.0000$ (0.0017, 0.0043) | $p = 1.0000$ (−0.0155, −0.0094) |
| | GRU-BS + Trans. Reranker | $p = 1.0000$ (−0.0253, −0.0139) | $p = 1.0000$ (0.0276, 0.0475) | $p = 1.0000$ (0.0065, 0.0111) | $p = 1.0000$ (0.0020, 0.0045) | $p = 1.0000$ (−0.0152, −0.0091) |
| GRU-BS + GRU Reranker | GRU (Meloni et al., 2021) | $p < 0.0001^*$ (0.0312, 0.0442)* | $p < 0.0001^*$ (−0.0853, −0.0631)* | $p < 0.0001^*$ (−0.0200, −0.0148)* | $p < 0.0001^*$ (−0.0074, −0.0047)* | $p < 0.0001^*$ (0.0196, 0.0268)* |
| | Trans (Kim et al., 2023) | $p < 0.0001^*$ (0.0216, 0.0354)* | $p < 0.0001^*$ (−0.0629, −0.0380)* | $p < 0.0001^*$ (−0.0147, −0.0089)* | $p < 0.0001^*$ (−0.0049, −0.0018)* | $p < 0.0001^*$ (0.0128, 0.0204)* |
| | GRU-BS | $p < 0.0001^*$ (0.0141, 0.0249)* | $p < 0.0001^*$ (−0.0489, −0.0282)* | $p < 0.0001^*$ (−0.0114, −0.0066)* | $p < 0.0001^*$ (−0.0042, −0.0017)* | $p < 0.0001^*$ (0.0093, 0.0153)* |
| | GRU-BS + Trans. Reranker | $p = 0.3882$ (−0.0058, 0.0058) | $p = 0.3625$ (−0.0122, 0.0097) | $p = 0.3625$ (−0.0029, 0.0023) | $p = 0.5804$ (−0.0011, 0.0017) | $p = 0.3727$ (−0.0030, 0.0035) |
| GRU-BS + Trans. Reranker | GRU (Meloni et al., 2021) | $p < 0.0001^*$ (0.0311, 0.0444)* | $p < 0.0001^*$ (−0.0830, −0.0620)* | $p < 0.0001^*$ (−0.0194, −0.0145)* | $p < 0.0001^*$ (−0.0077, −0.0050)* | $p < 0.0001^*$ (0.0193, 0.0264)* |
| | Trans (Kim et al., 2023) | $p < 0.0001^*$ (0.0214, 0.0355)* | $p < 0.0001^*$ (−0.0604, −0.0366)* | $p < 0.0001^*$ (−0.0142, −0.0086)* | $p < 0.0001^*$ (−0.0052, −0.0021)* | $p < 0.0001^*$ (0.0125, 0.0200)* |
| | GRU-BS | $p < 0.0001^*$ (0.0137, 0.0252)* | $p < 0.0001^*$ (−0.0468, −0.0275)* | $p < 0.0001^*$ (−0.0110, −0.0064)* | $p < 0.0001^*$ (−0.0045, −0.0020)* | $p < 0.0001^*$ (0.0091, 0.0150)* |
| | GRU-BS + GRU Reranker | $p = 0.6118$ (−0.0058, 0.0058) | $p = 0.6375$ (−0.0092, 0.0130) | $p = 0.6375$ (−0.0022, 0.0030) | $p = 0.4196$ (−0.0017, 0.0011) | $p = 0.6273$ (−0.0036, 0.0028) |

Table 27: Reconstruction significance test results for WikiHan-aug. Asterisks indicates that Reconstruction System 1 performs better than Reconstruction System 2 with the corresponding test (*p*-value or CI).

| Reconstruction System 1 | Reconstruction System 2 | ACC%↑ | TED↓ | TER↓ | FER↓ | BCFS↑ |
|---|---|---|---|---|---|---|
| GRU (Meloni et al., 2021) | Trans (Kim et al., 2023) | $p = 1.0000$ (−0.0658, −0.0224) | $p = 1.0000$ (0.0571, 0.1453) | $p = 1.0000$ (0.0139, 0.0348) | $p = 0.9867$ (−0.0015, 0.0084) | $p = 1.0000$ (−0.0369, −0.0146) |
| | GRU-BS | $p = 0.9991$ (−0.0516, −0.0059) | $p = 0.9914$ (−0.0009, 0.1053) | $p = 0.9975$ (0.0026, 0.0269) | $p = 0.6772$ (−0.0056, 0.0038) | $p = 1.0000$ (−0.0417, −0.0132) |
| | GRU-BS + GRU Reranker | $p = 1.0000$ (−0.0813, −0.0370) | $p = 1.0000$ (0.0739, 0.1652) | $p = 1.0000$ (0.0193, 0.0404) | $p = 0.9953$ (0.0000, 0.0087) | $p = 1.0000$ (−0.0557, −0.0309) |
| | GRU-BS + Trans. Reranker | $p = 1.0000$ (−0.0988, −0.0540) | $p = 1.0000$ (0.0935, 0.1891) | $p = 1.0000$ (0.0237, 0.0455) | $p = 0.9990$ (0.0020, 0.0111) | $p = 1.0000$ (−0.0600, −0.0349) |
| Trans (Kim et al., 2023) | GRU (Meloni et al., 2021) | $p < 0.0001$* (0.0230, 0.0658)* | $p < 0.0001$* (−0.1463, −0.0571)* | $p < 0.0001$* (−0.0348, −0.0140)* | $p = 0.0133$ (−0.0083, 0.0016) | $p < 0.0001$* (0.0143, 0.0370)* |
| | GRU-BS | $p = 0.0468$ (−0.0087, 0.0405) | $p = 0.0142$ (−0.0935, −0.0028)* | $p = 0.0199$ (−0.0199, 0.0011) | $p = 0.0199$ (−0.0099, 0.0012) | $p = 0.7331$ (−0.0137, 0.0101) |
| | GRU-BS + GRU Reranker | $p = 0.9119$ (−0.0398, 0.0096) | $p = 0.8416$ (−0.0186, 0.0562) | $p = 0.9029$ (−0.0032, 0.0143) | $p = 0.5751$ (−0.0043, 0.0063) | $p = 0.9998$ (−0.0272, −0.0081) |
| | GRU-BS + Trans. Reranker | $p = 0.9969$ (−0.0562, −0.0086) | $p = 0.9920$ (0.0031, 0.0804) | $p = 0.9907$ (0.0017, 0.0197) | $p = 0.8543$ (−0.0024, 0.0083) | $p = 1.0000$ (−0.0320, −0.0126) |
| GRU-BS | GRU (Meloni et al., 2021) | $p < 0.0010$* (0.0056, 0.0516)* | $p = 0.0086$* (−0.1048, −0.0027)* | $p = 0.0025$* (−0.0268, −0.0034)* | $p = 0.3228$ (−0.0039, 0.0054) | $p < 0.0001$* (0.0141, 0.0413)* |
| | Trans (Kim et al., 2023) | $p = 0.9532$ (−0.0410, 0.0084) | $p = 0.9858$ (0.0025, 0.0944) | $p = 0.9801$ (−0.0011, 0.0202) | $p = 0.9801$ (−0.0014, 0.0096) | $p = 0.2669$ (−0.0097, 0.0135) |
| | GRU-BS + GRU Reranker | $p = 0.9963$ (−0.0571, −0.0059) | $p = 0.9985$ (0.0211, 0.1130) | $p = 0.9985$ (0.0047, 0.0253) | $p = 0.9923$ (0.0002, 0.0103) | $p = 0.9945$ (−0.0284, −0.0034) |
| | GRU-BS + Trans. Reranker | $p = 1.0000$ (−0.0736, −0.0236) | $p = 0.9999$ (0.0416, 0.1381) | $p = 0.9999$ (0.0094, 0.0310) | $p = 0.9976$ (0.0019, 0.0124) | $p = 0.9996$ (−0.0330, −0.0076) |
| GRU-BS + GRU Reranker | GRU (Meloni et al., 2021) | $p < 0.0001$* (0.0370, 0.0817)* | $p < 0.0001$* (−0.1638, −0.0743)* | $p < 0.0001$* (−0.0402, −0.0194)* | $p = 0.0047$* (−0.0087, 0.0002) | $p < 0.0001$* (0.0313, 0.0553)* |
| | Trans (Kim et al., 2023) | $p = 0.0881$ (−0.0093, 0.0385) | $p = 0.1584$ (−0.0553, 0.0186) | $p = 0.0971$ (−0.0141, 0.0032) | $p = 0.4249$ (−0.0063, 0.0042) | $p < 0.0010$* (0.0080, 0.0272)* |
| | GRU-BS | $p = 0.0037$* (0.0042, 0.0556)* | $p = 0.0015$* (−0.1130, −0.0189)* | $p = 0.0015$* (−0.0253, −0.0042)* | $p = 0.0077$* (−0.0104, −0.0000)* | $p = 0.0055$* (0.0028, 0.0282)* |
| | GRU-BS + Trans. Reranker | $p = 0.9029$ (−0.0432, 0.0067) | $p = 0.8663$ (−0.0169, 0.0639) | $p = 0.8663$ (−0.0038, 0.0144) | $p = 0.8029$ (−0.0028, 0.0071) | $p = 0.7915$ (−0.0158, 0.0059) |
| GRU-BS + Trans. Reranker | GRU (Meloni et al., 2021) | $p < 0.0001$* (0.0548, 0.0978)* | $p < 0.0001$* (−0.1883, −0.0950)* | $p < 0.0001$* (−0.0455, −0.0241)* | $p = 0.0010$* (−0.0110, −0.0018)* | $p < 0.0001$* (0.0354, 0.0597)* |
| | Trans (Kim et al., 2023) | $p = 0.0031$* (0.0081, 0.0559)* | $p = 0.0080$* (−0.0801, −0.0021)* | $p = 0.0093$* (−0.0199, −0.0015)* | $p = 0.1457$ (−0.0083, 0.0025) | $p < 0.0001$* (0.0123, 0.0321)* |
| | GRU-BS | $p < 0.0001$* (0.0224, 0.0724)* | $p < 0.0010$* (−0.1373, −0.0404)* | $p < 0.0010$* (−0.0308, −0.0091)* | $p = 0.0024$* (−0.0125, −0.0019)* | $p < 0.0010$* (0.0067, 0.0328)* |
| | GRU-BS + GRU Reranker | $p = 0.0971$ (−0.0084, 0.0404) | $p = 0.1337$ (−0.0630, 0.0185) | $p = 0.1337$ (−0.0141, 0.0041) | $p = 0.1971$ (−0.0068, 0.0029) | $p = 0.2085$ (−0.0066, 0.0152) |

Table 28: Reconstruction significance test results for Hóu. Asterisks indicates that Reconstruction System 1 performs better than Reconstruction System 2 with the corresponding test (*p*-value or CI).

| Reconstruction System 1 | Reconstruction System 2 | ACC%↑ | TED↓ | TER↓ | FER↓ | BCFS↑ |
|---|---|---|---|---|---|---|
| GRU (Meloni et al., 2021) | Trans (Kim et al., 2023) | $p = 0.9999$ (−0.0174, −0.0052) | $p = 1.0000$ (0.0567, 0.0886) | $p = 1.0000$ (0.0077, 0.0116) | $p = 0.9993$ (0.0005, 0.0021) | $p = 1.0000$ (−0.0167, −0.0119) |
| | GRU-BS | $p = 0.9979$ (−0.0124, −0.0013) | $p = 1.0000$ (0.0487, 0.0810) | $p = 1.0000$ (0.0206, 0.0246) | $p = 1.0000$ (0.0028, 0.0045) | $p = 1.0000$ (−0.0154, −0.0100) |
| | GRU-BS + GRU Reranker | $p = 1.0000$ (−0.0260, −0.0142) | $p = 1.0000$ (0.0846, 0.1159) | $p = 1.0000$ (0.0246, 0.0285) | $p = 1.0000$ (0.0046, 0.0062) | $p = 1.0000$ (−0.0210, −0.0160) |
| | GRU-BS + Trans. Reranker | $p = 1.0000$ (−0.0252, −0.0131) | $p = 1.0000$ (0.0843, 0.1167) | $p = 1.0000$ (0.0246, 0.0286) | $p = 1.0000$ (0.0049, 0.0065) | $p = 1.0000$ (−0.0212, −0.0160) |
| Trans (Kim et al., 2023) | GRU (Meloni et al., 2021) | $p < 0.0001$* (0.0053, 0.0174)* | $p < 0.0001$* (−0.0890, −0.0570)* | $p < 0.0010$* (−0.0116, −0.0077)* | $p < 0.0001$* (−0.0021, −0.0004)* | $p < 0.0001$* (0.0119, 0.0167)* |
| | GRU-BS | $p = 0.0881$ (−0.0017, 0.0103) | $p = 0.1280$ (−0.0222, 0.0055) | $p = 1.0000$ (0.0114, 0.0145) | $p = 1.0000$ (0.0017, 0.0031) | $p = 0.0925$ (−0.0007, 0.0039) |
| | GRU-BS + GRU Reranker | $p = 0.9992$ (−0.0153, −0.0025) | $p = 1.0000$ (0.0141, 0.0408) | $p = 1.0000$ (0.0155, 0.0184) | $p = 1.0000$ (0.0035, 0.0048) | $p = 1.0000$ (−0.0063, −0.0021) |
| | GRU-BS + Trans. Reranker | $p = 0.9973$ (−0.0144, −0.0015) | $p = 1.0000$ (0.0136, 0.0414) | $p = 1.0000$ (0.0154, 0.0185) | $p = 1.0000$ (0.0038, 0.0051) | $p = 1.0000$ (−0.0065, −0.0020) |
| GRU-BS | GRU (Meloni et al., 2021) | $p = 0.0021$* (0.0017, 0.0123)* | $p < 0.0001$* (−0.0811, −0.0490)* | $p < 0.0001$* (−0.0246, −0.0206)* | $p < 0.0001$* (−0.0045, −0.0028)* | $p < 0.0001$* (0.0100, 0.0154)* |
| | Trans (Kim et al., 2023) | $p = 0.9119$ (−0.0102, 0.0017) | $p = 0.8720$ (−0.0057, 0.0216) | $p < 0.0001$* (−0.0145, −0.0115)* | $p < 0.0001$* (−0.0031, −0.0017)* | $p = 0.9075$ (−0.0038, 0.0007) |
| | GRU-BS + GRU Reranker | $p = 1.0000$ (−0.0189, −0.0075) | $p = 1.0000$ (0.0218, 0.0485) | $p = 1.0000$ (0.0024, 0.0054) | $p = 1.0000$ (0.0011, 0.0024) | $p = 1.0000$ (−0.0082, −0.0034) |
| | GRU-BS + Trans. Reranker | $p = 1.0000$ (−0.0182, −0.0063) | $p = 1.0000$ (0.0212, 0.0495) | $p = 1.0000$ (0.0024, 0.0055) | $p = 1.0000$ (0.0013, 0.0027) | $p = 1.0000$ (−0.0083, −0.0034) |
| GRU-BS + GRU Reranker | GRU (Meloni et al., 2021) | $p < 0.0001$* (0.0146, 0.0260)* | $p < 0.0001$* (−0.1160, −0.0845)* | $p < 0.0001$* (−0.0285, −0.0246)* | $p < 0.0001$* (−0.0062, −0.0046)* | $p < 0.0001$* (0.0160, 0.0211)* |
| | Trans (Kim et al., 2023) | $p < 0.0010$* (0.0027, 0.0154)* | $p < 0.0001$* (−0.0410, −0.0141)* | $p < 0.0001$* (−0.0184, −0.0155)* | $p < 0.0001$* (−0.0048, −0.0035)* | $p < 0.0001$* (0.0021, 0.0064)* |
| | GRU-BS | $p < 0.0001$* (0.0076, 0.0191)* | $p < 0.0001$* (−0.0490, −0.0219)* | $p < 0.0001$* (−0.0055, −0.0024)* | $p < 0.0001$* (−0.0024, −0.0011)* | $p < 0.0001$* (0.0035, 0.0083)* |
| | GRU-BS + Trans. Reranker | $p = 0.3474$ (−0.0053, 0.0072) | $p = 0.6066$ (−0.0133, 0.0137) | $p = 0.6066$ (−0.0015, 0.0015) | $p = 0.8830$ (−0.0003, 0.0009) | $p = 0.6066$ (−0.0024, 0.0023) |
| GRU-BS + Trans. Reranker | GRU (Meloni et al., 2021) | $p < 0.0001$* (0.0136, 0.0252)* | $p < 0.0001$* (−0.1165, −0.0843)* | $p < 0.0001$* (−0.0286, −0.0246)* | $p < 0.0001$* (−0.0065, −0.0049)* | $p < 0.0001$* (0.0160, 0.0211)* |
| | Trans (Kim et al., 2023) | $p = 0.0027$* (0.0014, 0.0146)* | $p < 0.0001$* (−0.0417, −0.0138)* | $p < 0.0001$* (−0.0185, −0.0154)* | $p < 0.0001$* (−0.0051, −0.0038)* | $p < 0.0001$* (0.0021, 0.0065)* |
| | GRU-BS | $p < 0.0001$* (0.0065, 0.0182)* | $p < 0.0001$* (−0.0501, −0.0218)* | $p < 0.0001$* (−0.0056, −0.0024)* | $p < 0.0001$* (−0.0028, −0.0014)* | $p < 0.0001$* (0.0034, 0.0084)* |
| | GRU-BS + GRU Reranker | $p = 0.6526$ (−0.0071, 0.0052) | $p = 0.3934$ (−0.0139, 0.0132) | $p = 0.3934$ (−0.0016, 0.0015) | $p = 0.1170$ (−0.0009, 0.0003) | $p = 0.3934$ (−0.0022, 0.0024) |

Table 29: Reconstruction significance test results for Rom-phon. Asterisks indicates that Reconstruction System 1 performs better than Reconstruction System 2 with the corresponding test (*p*-value or CI).

| Reconstruction System 1 | Reconstruction System 2 | ACC%↑ | TED↓ | TER↓ | FER↓ | BCFS↑ |
|---|---|---|---|---|---|---|
| GRU (Meloni et al., 2021) | Trans (Kim et al., 2023) | $p = 1.0000$ (−0.0214, −0.0128) | $p = 1.0000$ (0.0249, 0.0495) | $p = 1.0000$ (0.0030, 0.0064) | - | $p = 1.0000$ (−0.0097, −0.0055) |
| | GRU-BS | $p = 1.0000$ (−0.0217, −0.0130) | $p = 1.0000$ (0.0379, 0.0595) | $p = 1.0000$ (0.0153, 0.0180) | - | $p = 1.0000$ (−0.0106, −0.0067) |
| | GRU-BS + GRU Reranker | $p = 1.0000$ (−0.0366, −0.0287) | $p = 1.0000$ (0.0682, 0.0882) | $p = 1.0000$ (0.0186, 0.0212) | - | $p = 1.0000$ (−0.0158, −0.0124) |
| | GRU-BS + Trans. Reranker | $p = 1.0000$ (−0.0353, −0.0274) | $p = 1.0000$ (0.0672, 0.0869) | $p = 1.0000$ (0.0185, 0.0210) | - | $p = 1.0000$ (−0.0157, −0.0122) |
| Trans (Kim et al., 2023) | GRU (Meloni et al., 2021) | $p < 0.0001$* (0.0128, 0.0216)* | $p < 0.0001$* (−0.0493, −0.0254)* | $p < 0.0001$* (−0.0064, −0.0031)* | - | $p < 0.0001$* (0.0055, 0.0097)* |
| | GRU-BS | $p = 0.7419$ (−0.0040, 0.0041) | $p = 0.9848$ (−0.0011, 0.0222) | $p = 1.0000$ (0.0103, 0.0133) | - | $p = 0.7591$ (−0.0030, 0.0011) |
| | GRU-BS + GRU Reranker | $p = 1.0000$ (−0.0189, −0.0115) | $p = 1.0000$ (0.0293, 0.0515) | $p = 1.0000$ (0.0137, 0.0165) | - | $p = 1.0000$ (−0.0083, −0.0045) |
| | GRU-BS + Trans. Reranker | $p = 1.0000$ (−0.0178, −0.0101) | $p = 1.0000$ (0.0281, 0.0506) | $p = 1.0000$ (0.0136, 0.0164) | - | $p = 1.0000$ (−0.0082, −0.0043) |
| GRU-BS | GRU (Meloni et al., 2021) | $p < 0.0001$* (0.0130, 0.0214)* | $p < 0.0001$* (−0.0590, −0.0380)* | $p < 0.0001$* (−0.0179, −0.0152)* | - | $p < 0.0001$* (0.0067, 0.0105)* |
| | Trans (Kim et al., 2023) | $p = 0.2581$ (−0.0041, 0.0043) | $p = 0.0152$ (−0.0226, 0.0008) | $p < 0.0001$* (−0.0133, −0.0103)* | - | $p = 0.2409$ (−0.0011, 0.0030) |
| | GRU-BS + GRU Reranker | $p = 1.0000$ (−0.0191, −0.0115) | $p = 1.0000$ (0.0201, 0.0389) | $p = 1.0000$ (0.0022, 0.0043) | - | $p = 1.0000$ (−0.0072, −0.0038) |
| | GRU-BS + Trans. Reranker | $p = 1.0000$ (−0.0180, −0.0102) | $p = 1.0000$ (0.0191, 0.0380) | $p = 1.0000$ (0.0021, 0.0042) | - | $p = 1.0000$ (−0.0071, −0.0036) |
| GRU-BS + GRU Reranker | GRU (Meloni et al., 2021) | $p < 0.0001$* (0.0286, 0.0365)* | $p < 0.0001$* (−0.0882, −0.0684)* | $p < 0.0001$* (−0.0212, −0.0186)* | - | $p < 0.0001$* (0.0124, 0.0159)* |
| | Trans (Kim et al., 2023) | $p < 0.0001$* (0.0116, 0.0190)* | $p < 0.0001$* (−0.0517, −0.0295)* | $p < 0.0001$* (−0.0166, −0.0137)* | - | $p < 0.0001$* (0.0045, 0.0084)* |
| | GRU-BS | $p < 0.0001$* (0.0115, 0.0190)* | $p < 0.0001$* (−0.0390, −0.0199)* | $p < 0.0001$* (−0.0043, −0.0022)* | - | $p < 0.0001$* (0.0037, 0.0072)* |
| | GRU-BS + Trans. Reranker | $p = 0.3779$ (−0.0022, 0.0047) | $p = 0.4409$ (−0.0097, 0.0079) | $p = 0.4409$ (−0.0011, 0.0009) | - | $p = 0.5431$ (−0.0014, 0.0017) |
| GRU-BS + Trans. Reranker | GRU (Meloni et al., 2021) | $p < 0.0001$* (0.0274, 0.0353)* | $p < 0.0001$* (−0.0875, −0.0674)* | $p < 0.0001$* (−0.0211, −0.0185)* | - | $p < 0.0001$* (0.0122, 0.0158)* |
| | Trans (Kim et al., 2023) | $p < 0.0001$* (0.0103, 0.0180)* | $p < 0.0001$* (−0.0509, −0.0282)* | $p < 0.0001$* (−0.0165, −0.0136)* | - | $p < 0.0001$* (0.0044, 0.0083)* |
| | GRU-BS | $p < 0.0001$* (0.0102, 0.0180)* | $p < 0.0001$* (−0.0380, −0.0189)* | $p < 0.0001$* (−0.0042, −0.0021)* | - | $p < 0.0001$* (0.0036, 0.0071)* |
| | GRU-BS + GRU Reranker | $p = 0.6221$ (−0.0047, 0.0022) | $p = 0.5591$ (−0.0077, 0.0100) | $p = 0.5591$ (−0.0009, 0.0011) | - | $p = 0.4569$ (−0.0018, 0.0014) |

Table 30: Reconstruction significance test results for Rom-orth. Asterisks indicates that Reconstruction System 1 performs better than Reconstruction System 2 with the corresponding test (*p*-value or CI).

| Beam Search | | | Reflex Prediction (based on protoform candidates) | | | | | | | | Reranking Result | | |
|---|---|---|---|---|---|---|---|---|---|---|---|---|---|
| rank | $\hat{p}_i^{bs}$ | $m_i$ | Cantonese | Hakka | Jin | Mandarin | Hokkien | Wu | Xiang | $r_i$ | rank | $\hat{p}_i^{rk}$ | $s_i$ |
| 0 | kjwet入 | -0.2299 | ky:t˦ | kiet˦ | **tɕyə?˩** | tɕyɤ˧ | kuat˦ | **tɕyɪ?˧** | tɕie˦ | 0.2857 | 0 | **kwit入** | 0.3318 |
| 1 | kit入 | -0.2866 | ket˧ | **kit˦** | tɕiə?˩ | tɕi˧ | kit˦ | tɕiɪ?˧ | tɕi˦ | 0.1429 | 1 | kjwet入 | 0.1301 |
| 2 | **kwit入** | -0.3882 | **kʷet˧** | kiut˦ | **tɕyə?˩** | tɕyɤ˧ | kut˦ | **tɕyɪ?˧** | **tɕy˦** | 0.5714 | 2 | kjiwt入 | -0.0567 |
| 3 | kjit入 | -0.3979 | ket˧ | **kit˦** | tɕiə?˩ | tɕi˧ | kit˦ | tɕiɪ?˧ | tɕi˦ | 0.1429 | 3 | kit入 | -0.1066 |
| 4 | kjiwt入 | -0.5967 | **kʷet˧** | kiut˦ | **tɕyə?˩** | tɕyɤ˧ | kut˦ | **tɕyɪ?˧** | tɕyn˧ | 0.4286 | 4 | kjit入 | -0.2179 |
| 5 | kjet入 | -0.7380 | ki:t˦ | kiet˦ | tɕiə?˩ | tɕie˧ | kiet˦ | tɕiɪ?˧ | tɕie˦ | 0.1429 | 5 | kjet入 | -0.5580 |
| 6 | kɛt入 | -0.8579 | ka:t˦ | kat˦ | tɕiə?˩ | tɕie˧ | kat˦ | kaʔ˧ | kɤ˦ | 0.0000 | 6 | kɛt入 | -0.8579 |
|  |  |  | **kʷet˧** | **kit˦** | **tɕyə?˩** | **tɕy˧** | **kiet˦** | **tɕyɪ?˧** | **tɕy˦** | - |  |  |  |

Figure 4: Successful reranking of 橘 *kwit*入 'mandarin orange'.

| Beam Search | | | Reflex Prediction (based on protoform candidates) | | | | | | Reranking Result | | |
|---|---|---|---|---|---|---|---|---|---|---|---|
| rank | $\hat{p}_i^{bs}$ | $m_i$ | Cantonese | Hakka | Mandarin | Hokkien | Wu | $r_i$ | rank | $\hat{p}_i^{rk}$ | $s_i$ |
| 0 | kwaj去 | -0.2234 | kʷɔ:˦ | kui˧ | kʰuai˥ | kue˨ | kuɛ˦ | 0.0000 | 0 | **ɣwaj去** | 0.3959 |
| 1 | ʔwaj去 | -0.3368 | wu:y˦ | voi˧ | uei˥ | ue˨ | uɛ˦ | 0.0000 | 1 | ɣwoj去 | 0.2032 |
| 2 | **ɣwaj去** | -0.3601 | wu:y˦ | **fi˧** | **xuei˥** | ue˨ | **ɦuɛ˦** | 0.6000 | 2 | kwaj去 | -0.2234 |
| 3 | kwoj去 | -0.3787 | kʷu:y˦ | kui˧ | kuei˥ | kue˨ | kuɛ˦ | 0.0000 | 3 | ʔwaj去 | -0.3368 |
| 4 | ɣwoj去 | -0.5528 | wu:y˦ | **fi˧** | **xuei˥** | hue˨ | **ɦuɛ˦** | 0.6000 | 4 | kwoj去 | -0.3787 |
| 5 | ʔwoj去 | -0.5998 | wu:y˦ | ve˧ | uei˥ | ue˨ | uɛ˦ | 0.0000 | 5 | ʔwoj去 | -0.5998 |
| 6 | kʰwaj去 | -0.8277 | fu:y˦ | kʰuai˧ | kʰuai˥ | kʰue˨ | kʰuɛ˦ | 0.0000 | 6 | kʰwaj去 | -0.8277 |
|  |  |  | **kʰu:y˦** | **fi˧** | **xuei˥** | **kue˨** | **ɦuɛ˦** | - |  |  |  |

Figure 5: Successful reranking of 繪 *ɣwaj*去 'to draw'.

| Beam Search | | | Reflex Prediction (based on protoform candidates) | | | | Reranking Result | | |
|---|---|---|---|---|---|---|---|---|---|
| rank | $\hat{p}_i^{bs}$ | $m_i$ | Cantonese | Mandarin | Hokkien | $r_i$ | rank | $\hat{p}_i^{rk}$ | $s_i$ |
| 0 | **sew平** | -0.1557 | si:u˧ | ɕiau˧ | ɕiau˧ | 0.0000 | 0 | suw平 | 0.0196 |
| 1 | suw平 | -0.4004 | seu˧ | **soɤ˧** | sau˧ | 0.3333 | 1 | suw上 | -0.0471 |
| 2 | suw上 | -0.4671 | **seu˨** | soɤ˨ | sau˨ | 0.3333 | 2 | **sew平** | -0.1557 |
| 3 | sju平 | -0.5906 | søy˧ | ɕy˧ | ɕi˧ | 0.0000 | 3 | sju平 | -0.5906 |
| 4 | sew上 | -0.5909 | si:u˨ | ɕiau˨ | ɕiau˨ | 0.0000 | 4 | sew上 | -0.5909 |
| 5 | sju上 | -0.9313 | søy˨ | ɕy˨ | ɕi˨ | 0.0000 | 5 | sju上 | -0.9313 |
|  |  |  | **seu˨** | **soɤ˧** | **sɤ˨** | - |  |  |  |

Figure 6: Unsuccessful reranking of 艘 *sew*平 'small boat'.

| Beam Search | | | Reflex Prediction (based on protoform candidates) | | | | | | | | Reranking Result | | |
|---|---|---|---|---|---|---|---|---|---|---|---|---|---|
| rank | $\hat{p}_i^{bs}$ | $m_i$ | Cantonese | Gan | Hakka | Jin | Mandarin | Hokkien | Wu | Xiang | $r_i$ | rank | $\hat{p}_i^{rk}$ | $s_i$ |
| 0 | kjun平 | -0.1272 | kən˧ | **tɕyn˨** | **kiun˦** | **tɕyŋ˧** | **tɕyn˧** | **kun˧** | tɕyŋ˥ | tɕyn˦ | 0.8750 | 0 | kwin平 | 1.0834 |
| 1 | kwin平 | -0.1766 | **kʷen˧** | **tɕyn˨** | **kiun˦** | **tɕyŋ˧** | **tɕyn˧** | **kun˧** | **tɕyŋ˥** | **tɕyn˦** | 1.0000 | 1 | **kjun平** | 0.9753 |
| 2 | kin平 | -0.6991 | ken˧ | tɕin˨ | kin˦ | tɕiŋ˧ | tɕin˧ | kin˧ | tɕɪŋ˥ | tɕin˦ | 0.0000 | 2 | kin平 | -0.6991 |
| 3 | kwen平 | -0.9995 | ky:n˧ | tɕyen˨ | kien˦ | tɕye˧ | tɕyan˧ | kuan˧ | tɕyø˥ | tɕye˦ | 0.0000 | 3 | kwen平 | -0.9995 |
| 4 | kjin平 | -1.0021 | ken˧ | tɕin˨ | kin˦ | tɕiŋ˧ | tɕin˧ | kin˧ | tɕɪŋ˥ | tɕin˦ | 0.0000 | 4 | kjin平 | -1.0021 |
| 5 | kjon平 | -1.2847 | ki:n˧ | tɕien˨ | kien˦ | tɕie˧ | tɕien˧ | kien˧ | tɕɪŋ˥ | tɕiẽ˦ | 0.0000 | 5 | kjon平 | -1.2847 |
|  |  |  | **kʷen˧** | **tɕyn˨** | **kiun˦** | **tɕyŋ˧** | **tɕyn˧** | **kun˧** | **tɕyŋ˥** | **tɕyn˦** | - |  |  |  |

Figure 7: Unsuccessful reranking of 君 *kjun*平 'sovereign'.

| Beam Search | | |
|---|---|---|
| rank | $\hat{p}_i^{bs}$ | $m_i$ |
| 0 | dıskrɛptıam | -0.0876 |
| 1 | **dıskrɛpantıam** | -0.1073 |
| 2 | dıskrɛpantıam | -0.3963 |
| 3 | dıskrɛpsıam | -0.4401 |
| 4 | dıskrɛpsantıam | -0.4729 |

| Reflex Prediction (based on protoform candidates) | | | |
|---|---|---|---|
| Italian | Spanish | Portuguese | $r_i$ |
| diskrɛtsa | diskrepθja | diʃkɹepsje | 0.0000 |
| **diskrepantsa** | **diskrepanθja** | **diʃkɹɨpẽŋsje** | 1.0000 |
| **diskrepantsa** | **diskrepanθja** | **diʃkɹɨpẽŋsje** | 1.0000 |
| diskressia | diskrepsja | diʃkɹɨpsie | 0.0000 |
| diskressantsa | diskrepsanθja | diʃkɹɨpsẽŋsje | 0.0000 |
| **diskrepantsa** | **diskrepanθja** | **diʃkɹɨpẽŋsje** | - |

| Reranking Result | | |
|---|---|---|
| rank | $\hat{p}_i^{rk}$ | $s_i$ |
| 0 | **dıskrɛpantıam** | 0.4777 |
| 1 | dıskrepantıam | 0.1887 |
| 2 | dıskrɛptıam | -0.0876 |
| 3 | dıskrɛpsıam | -0.4401 |
| 4 | dıskrɛpsantıam | -0.4729 |

Figure 8: Succssful reranking of *discrepantiam* 'discordance'.

| Beam Search | | |
|---|---|---|
| rank | $\hat{p}_i^{bs}$ | $m_i$ |
| 0 | ɛksɛkwikɛrɛ | -0.2887 |
| 1 | ɛksɛkwɪtarɛ | -0.3491 |
| 2 | ɛkssɛkwɪtarɛ | -0.3907 |
| 3 | ɛksɛkwirɛ | -0.4299 |
| 4 | **ɛkssɛkwi** | -0.4323 |
| 5 | ɛksɛkwi | -0.4541 |

| Reflex Prediction (based on protoform candidates) | | | |
|---|---|---|---|
| French | Italian | Spanish | $r_i$ |
| ɛgzeke | ezekire | eksejir | 0.0000 |
| ɛgzekite | ezekwitare | eksekitar | 0.0000 |
| ɛgzekite | ezekwitare | eksekiðar | 0.0000 |
| ɛgzeke | ezekwire | eksejir | 0.0000 |
| **ɛgzekyte** | **ezegwire** | eksejir | 0.6667 |
| **ɛgzekyte** | **ezegwire** | eksejir | 0.6667 |
| **ɛgzekyte** | **ezegwire** | exekutar | - |

| Reranking Result | | |
|---|---|---|
| rank | $\hat{p}_i^{rk}$ | $s_i$ |
| 0 | **ɛkssɛkwi** | -0.0423 |
| 1 | ɛksɛkwi | -0.0641 |
| 2 | ɛksɛkwikɛrɛ | -0.2887 |
| 3 | ɛksɛkwɪtarɛ | -0.3491 |
| 4 | ɛkssɛkwɪtarɛ | -0.3907 |
| 5 | ɛksɛkwirɛ | -0.4299 |

Figure 9: Successful reranking of *exsequi* 'follow after'.

| Beam Search | | |
|---|---|---|
| rank | $\hat{p}_i^{bs}$ | $m_i$ |
| 0 | **prokrastınarɛ** | -0.1000 |
| 1 | prokrastinarɛ | -0.1154 |
| 2 | prokrastınari | -0.2423 |
| 3 | prokrastinari | -0.2629 |
| 4 | prɔkrastinarɛ | -0.6461 |
| 5 | prɔkrastınarɛ | -0.6774 |

| Reflex Prediction (based on protoform candidates) | | | | |
|---|---|---|---|---|
| French | Italian | Spanish | Portuguese | $r_i$ |
| pʁɔkʁastine | prokrastinare | prokrastinar | pɹukɹaʃtinaɹ | 0.7500 |
| **pʁɔkʁastine** | **prokrastinare** | **prokrastinar** | **pɹukɹeʃtinaɹ** | 1.0000 |
| pʁɔkʁastine | prokrastinare | prokrastinar | pɹukɹaʃtinaɹ | 0.7500 |
| **pʁɔkʁastine** | **prokrastinare** | **prokrastinar** | **pɹukɹeʃtinaɹ** | 1.0000 |
| **pʁɔkʁastine** | **prokrastinare** | **prokrastinar** | **pɹukɹeʃtinaɹ** | 1.0000 |
| pʁɔkʁastine | prokrastinare | prokrastinar | pɹukɹaʃtinaɹ | 0.7500 |
| pʁɔkʁastine | prokrastinare | prokrastinar | pɹukɹeʃtinaɹ | - |

| Reranking Result | | |
|---|---|---|
| rank | $\hat{p}_i^{rk}$ | $s_i$ |
| 0 | prokrastinarɛ | 0.4696 |
| 1 | **prokrastınarɛ** | 0.3388 |
| 2 | prokrastinari | 0.3221 |
| 3 | prokrastınari | 0.1965 |
| 4 | prɔkrastinarɛ | -0.0611 |
| 5 | prɔkrastınarɛ | -0.2386 |

Figure 10: Unsuccessful reranking of *procrastinare* 'to defer'.

| Beam Search | | |
|---|---|---|
| rank | $\hat{p}_i^{bs}$ | $m_i$ |
| 0 | **aɪkwɪnɔktɪʊm** | -0.1209 |
| 1 | ɛkwɪnɔktɪʊm | -0.1384 |
| 2 | aɪkwɪnoktɪʊm | -0.2969 |
| 3 | ɛkwɪnoktɪʊm | -0.3124 |
| 4 | aɪkwɪnoktɪʊm | -0.4513 |
| 5 | ɛkwɪnɔktɪʊm | -0.4936 |

| Reflex Prediction (based on protoform candidates) | | | | | |
|---|---|---|---|---|---|
| Romanian | French | Italian | Spanish | Portuguese | $r_i$ |
| ekinokts | ekinoks | ekwinɔtso | **ekinokθjo** | ekinusjʊ | 0.2000 |
| **ekinoktsiw** | **ekinɔks** | **ekwinɔtsio** | **ekinokθjo** | ekinɔsjʊ | 0.8000 |
| ekinokts | ekinoks | ekwinɔtso | **ekinokθjo** | ekinɔsjʊ | 0.2000 |
| **ekinoktsiw** | **ekinɔks** | **ekwinɔtsio** | **ekinokθjo** | ekinɔsjʊ | 0.8000 |
| ekinokts | ekinoks | ekwinoktsio | **ekinokθjo** | ekinusjʊ | 0.2000 |
| **ekinoktsiw** | **ekinɔks** | **ekwinɔtsio** | **ekinokθjo** | ekinɔsjʊ | 0.8000 |
| ekinoktsiw | ekinɔks | ekwinɔtsio | ekinokθjo | ekwinɔsjʊ | - |

| Reranking Result | | |
|---|---|---|
| rank | $\hat{p}_i^{rk}$ | $s_i$ |
| 0 | ɛkwɪnɔktɪʊm | 0.3296 |
| 1 | ɛkwɪnɔktɪʊm | 0.1556 |
| 2 | **aɪkwɪnɔktɪʊm** | -0.0039 |
| 3 | ɛkwɪnɔktɪʊm | -0.0256 |
| 4 | aɪkwɪnɔktɪʊm | -0.1799 |
| 5 | aɪkwɪnɔktɪʊm | -0.3343 |

Figure 11: Unsuccessful reranking *aequinoctium* 'equinox'.